\documentclass{article}

     \usepackage[nonatbib,final]{neurips_2020}

\usepackage[utf8]{inputenc} 
\usepackage[T1]{fontenc}    
\usepackage{hyperref}       
\usepackage{url}            
\usepackage{booktabs}       
\usepackage{amsfonts}       
\usepackage{nicefrac}       
\usepackage{microtype}      
\usepackage{amsmath}
\usepackage{bm}
\usepackage{tabularx}
\usepackage{graphicx}
\usepackage[numbers]{natbib}
\usepackage{amssymb}
\usepackage{float}
\restylefloat{table}
\newcommand\citeinline[1]{%
  \citeauthor{#1}~[\citenum{#1}]}
\title{Optimal Mixture Weights for Off-Policy Evaluation with Multiple Behavior Policies}

\author{%
  Jinlin Lai\hspace{1em}Lixin Zou\hspace{1em}Jiaxing Song \\
  Department of Computer Science and Technology\\
  Tsinghua University\\
  \texttt{jinlinlai@cs.umass.edu,zoulx15@mails.tsinghua.edu.cn,jxsong@tsinghua.edu.cn} \\
}
\newtheorem{assumption}{Assumption}
\newtheorem{proposition}{Proposition}

\newtheorem{theorem}{Theorem}

\begin{document}

\maketitle

\begin{abstract}
	Off-policy evaluation is a key component of reinforcement learning which evaluates a target policy with offline data collected from behavior policies. It is a crucial step towards safe reinforcement learning and has been used in advertisement, recommender systems and many other applications. In these applications, sometimes the offline data is collected from multiple behavior policies. Previous works regard data from different behavior policies equally. Nevertheless, some behavior policies are better at producing good estimators while others are not. This paper starts with discussing how to correctly mix estimators produced by different behavior policies. We propose three ways to reduce the variance of the mixture estimator when all sub-estimators are unbiased or asymptotically unbiased. Furthermore, experiments on simulated recommender systems show that our methods are effective in reducing the Mean-Square Error of estimation.
\end{abstract}

\section{Introduction}
In applications of reinforcement learning \cite{Sutton1998ReinforcementLA}, it is usually unsafe or risky to use a policy without evaluating it. For example, in reinforcement learning based recommender systems, if a defective policy is deployed, it can cause irreversible loss like losing customers. To tackle the problem, Off-Policy Evaluation (OPE) algorithms are developed to evaluate a target policy with data collected from online behavior policies in an offline manner. OPE has been used to evaluate reinforcement learning applications in advertisements, recommender systems and many other areas \cite{Li2011UnbiasedOE,DBLP:journals/jmlr/BottouPCCCPRSS13,Joachims2016CounterfactualEA,Swaminathan2017OffpolicyEF,Gilotte2018OfflineAT}. The most influential algorithm in OPE is Doubly Robust estimation  \cite{DBLP:conf/icml/DudikLL11,Jiang2015DoublyRO}. Based on it, some recent works \cite{Thomas2016DataEfficientOP,Farajtabar2018MoreRD,BibautMVL19,DBLP:journals/corr/abs-1911-05811} in OPE explore different ways to reduce the Mean-Square Error (MSE) of estimation. However, few of them discuss deeply about how to evaluate with multiple behavior policies. In such cases, most current methods will directly go through after regarding data from different behavior policies as a whole. However, in later sections of this paper, we show that better results can be reached if we split data by behavior policy, construct split estimators for the split data and consider the mixture estimator of the split estimators. The root cause of this issue is that some behavior policies are "adding" high variance to the result. Therefore, if we can assign high weights to "good" behavior policies and low weights to "bad" ones, better estimation can be obtained. 

In this paper, we optimize the mixture weights for the mixture estimator by minimizing variance. This idea has been discussed in \citeinline{Aman2017EffectiveEva} for Importance Sampling estimators of contextual bandits. We generalize this idea to finite-horizon Markov Decision Process and derive naive mixture estimators for most OPE algorithms, including Importance Sampling, Weighted Importance Sampling, Doubly Robust estimation and Weighted Doubly Robust estimation. After exploiting the structure of reinforcement learning, we further propose mixture estimators and $\alpha\beta$ mixture estimators, which have theoretically lower variances than naive mixture estimators. To compute the optimal weights, we introduce Delta Method \cite{Owen2013MonteCarlo} from asymptotic statistics to estimate the variances and covariances of the components of weighted estimators. In our experiments on simulated recommender systems, we show that mixture estimators are effective in reducing the MSE of all the estimators.

\section{Preliminaries}
\subsection{Markov Decision Process}
Markov Decision Process (MDP) \cite{Sutton1998ReinforcementLA} is represented by $<\mathcal{S},\mathcal{A},R,P,P_0,\gamma>$, where $\mathcal{S}$ and $\mathcal{A}$ are state space and action space, $R(s,a)$ is a random variable indicating the immediate reward of taking action $a$ at state $s$, $P(\cdot|s,a)$ is the state transition distribution, $P_0(\cdot)$ is the distribution of initial state, and $\gamma$ is the discount factor. To interact in such environment, a policy $\pi$ is given and $\pi(a|s)$ is the probability of taking $a$ in state $s$. 
\subsection{Off-policy Evaluation with Single Behavior Policy}
\label{sec:22}
In the literature of reinforcement learning, there are many algorithms to evaluate a new policy $\pi$ with data collected from one behavioral policy $\pi_0$. In this section, we assume there are $N$ data trajectories and the $i$-th data is $(s_{i,0},a_{i,0},r_{i,0},s_{i,1},a_{i,1},r_{i,1},...)$. 

Direct Method (DM) fits $\mathbb{E}[R(s,a)]$ and $P(\cdot|s,a)$ by regression \cite{Jiang2015DoublyRO}. With the approximated functions $\hat{R}(s,a)$ and $\hat{P}(\cdot|s,a)$, the value functions are recursively updated ($\hat{V}_0(s)=0$):
\begin{gather}
\label{iteration}
	\hat{Q}_t(s,a)=\hat{R}(s,a)+\gamma\mathbb{E}_{s'\sim\hat{P}(s'|s,a)}[\hat{V}_{t-1}(s')]\quad \hat{V}_t(s)=\mathbb{E}_{a\sim \pi(a|s)}[\hat{Q}_t(s,a)],\\
	\hat{Q}(s,a)=\lim_{t\to\infty}\hat{Q}_t(s,a)\quad\hat{V}(s)=\lim_{t\to\infty}\hat{V}_t(s).
\end{gather}
The value of the new policy would be estimated by $\hat{V}_{DM}=\mathbb{E}_{s\sim P_0(s)}[\hat{V}(s)]$.

Besides DM, another family of OPE technique is Importance Sampling(IS). IS estimates the value by $\hat{V}_{IS}=\frac{1}{N}\sum_{i=1}^N\sum_{t=0}^{T}\gamma^t \rho_{i,t}r_{i,t}$, where $\rho_{i,t}=\prod_{\tau=0}^{t}\frac{\pi(a_{i,\tau}|s_{i,\tau})}{\pi_0(a_{i,\tau}|s_{i,\tau})}$ and $T$ is the maximal horizon of data.

DM typically has low variance and high bias. IS is proved to be unbiased but suffers from high variance. Doubly Robust estimation(DR) \cite{Jiang2015DoublyRO} combines DM and IS. It can be regarded as IS with control variates so it has lower variance than IS. We follow \citeinline{Thomas2016DataEfficientOP} and formulate DR as $\hat{V}_{DR}=\frac{1}{N}\sum_{i=1}^N\sum_{t=0}^{T}\gamma^t\left(\rho_{i,t-1}\hat{V}(s_{i,t})+\rho_{i,t}(r_{i,t}-\hat{Q}(s_{i,t},a_{i,t}))\right)$.

Weighted Importance Sampling(WIS) \cite{Powell1966WeightedUS} is also a variance reduction technique for IS. It is derived by replacing $\frac{\rho_{i,t}}{N}$ in IS with $w_{i,t}=\frac{\rho_{i,t}}{\sum_{i=1}^N \rho_{i,t}}$. WIS is asymptotically unbiased and has lower variance than IS. Similarly, Weighted Doubly Robust estimation(WDR) \cite{Thomas2016DataEfficientOP} has lower variance than DR. It also replaces $\frac{\rho_{i,t}}{N}$ in DR with $w_{i,t}$.

\subsection{Problem Setting}
In applications of reinforcement learning, there might be multiple behavior policies in the same environment. Suppose we have $M$ behavior policies $\pi_1,\pi_2,...,\pi_M$. The $i$-th behavior policy $\pi_i$ collects $n_i$ data. The $j$-th data from $\pi_i$ is $(s_{i,j,0},a_{i,j,0},r_{i,j,0},s_{i,j,1},a_{i,j,1},r_{i,j,1},...)$. With the data from $\pi_i$, we can build an asymptotically unbiased estimator $\hat{V}_i$ to evaluate the target policy $\pi$. This paper begins with constructing the mixture estimator of the $M$ estimators. The goal is to minimize the MSE of estimation. If we estimate $\theta$ with $\hat{\theta}$, the MSE is formulated as $\mathbb{E}[(\hat{\theta}-\theta)^2]=\mathbb{E}^2[\hat{\theta}-\theta]+\mathbb{V}[\hat{\theta}]$. For IS and DR, MSE reduces to variance so we directly minimize the variance. For WIS and WDR, by Delta Method \cite{Owen2013MonteCarlo}, the bias squared is $O(n^{-2})$ while the variance is $O(n^{-1})$. When $n$ is large, the bias squared is dominated by variance so we neglect the bias of weighted estimators and also directly minimize the variance. 

To make our idea clear, we define our symbol system of the following sections here. Bold letters are random variables (like $\bm{X}$). Letters with a subscript $j$ are samples of the corresponding random variable (for example, $X_{i,j}$ is a sample of $\bm{X}_i$). Letters with a hat are estimators (like $\hat{X}$). Letters with an arrow are vectors (like $\overrightarrow{X}$).

To guarantee the theoretical results in this paper, we give the following assumptions. 
\begin{assumption}
\label{ass:1}
	$\forall s,a,i$, if $\pi(a|s)>0$, then $\pi_{i}(a|s)>0$. Furthermore, there exists $\beta>0$, such that $\forall i,t$, $\rho_{i,t}\le \beta$.
\end{assumption}
\begin{assumption}
\label{ass:2}
	There exists $\zeta>0$, such that $\forall s,a$, $\forall r\sim R(s,a)$, $|r|\le \zeta$.
\end{assumption}
\begin{assumption}
\label{ass:3}
	For any estimator $\hat{\theta}$, we ignore its bias.
\end{assumption}
\begin{assumption}
\label{ass:4}
	Any two different data trajectories are independent.
\end{assumption}
\section{Optimal Mixture Weights with Multiple Behavior Policies}
\label{sec3}
\subsection{Value Estimators with Multiple Behavior Policies}
With multiple behavior policies, we can construct value estimators based on Section \ref{sec:22}. Define 
\begin{align}
	\rho_{i,j,t}&=\prod_{\tau=0}^t\frac{\pi(a_{i,j,\tau}|s_{i,j,\tau})}{\pi_i(a_{i,j,\tau}|s_{i,j,\tau})},\\
	w_{i,j,t}&=\frac{\rho_{i,j,t}}{\sum_{i'=1}^M\sum_{j'=1}^{n_{i'}}\rho_{i',j',t}},
\end{align}
then the IS, WIS, DR and WDR estimators are listed in Table \ref{tab:table1}.

\begin{table}
  \begin{center}
    \caption{Formulas for the vanilla estimators.}
    \label{tab:table1}
    \begin{tabular}{c|c} 
          Method&Estimator\\
      \hline
      IS & $\hat{V}_{IS}=\frac{1}{\sum_{i=1}^Mn_i}\sum_{i=1}^M\sum_{j=1}^{n_i}\sum_{t=0}^T\gamma^t\rho_{i,j,t}r_{i,j,t}$\\
      WIS & $\hat{V}_{WIS}=\sum_{i=1}^M\sum_{j=1}^{n_i}\sum_{t=0}^T\gamma^tw_{i,j,t}r_{i,j,t}$\\
      DR & $\hat{V}_{DR}=\frac{1}{\sum_{i=1}^Mn_i}\sum_{i=1}^M\sum_{j=1}^{n_i}\sum_{t=0}^T\gamma^t\left(\rho_{i,j,t-1}\hat{V}(s_{i,j,t})+\rho_{i,j,t}(r_{i,j,t}-\hat{Q}(s_{i,j,t},a_{i,j,t}))\right)$\\
      WDR & $\hat{V}_{WDR}=\sum_{i=1}^M\sum_{j=1}^{n_i}\sum_{t=0}^T\gamma^t\left(w_{i,j,t-1}\hat{V}(s_{i,j,t})+w_{i,j,t}(r_{i,j,t}-\hat{Q}(s_{i,j,t},a_{i,j,t}))\right)$\\
      SWIS & $\hat{V}_{SWIS}=\sum_{i=1}^M\frac{n_i}{\sum_{i'=1}^Mn_i'}\sum_{j=1}^{n_i}\sum_{t=0}^T\gamma^tu_{i,j,t}r_{i,j,t}$\\
      SWDR & $\hat{V}_{SWDR}=\sum_{i=1}^M\frac{n_i}{\sum_{i'=1}^Mn_i'}\sum_{j=1}^{n_i}\sum_{t=0}^T\gamma^t\left(u_{i,j,t-1}\hat{V}(s_{i,j,t})+u_{i,j,t}(r_{i,j,t}-\hat{Q}(s_{i,j,t},a_{i,j,t}))\right)$
     \end{tabular}
  \end{center}
\end{table}

Note that WIS and WDR normalize the importance weights across all data. We can also normalize inside each behavior policy by
\begin{gather}
	u_{i,j,t}=\frac{\rho_{i,j,t}}{\sum_{j'=1}^{n_{i'}}\rho_{i,j',t}}
\end{gather}
and construct split weighted estimators. We call them Split WIS (SWIS) and Split WDR (SWDR). See Table \ref{tab:table1} for their formulas.

In our experiments, we show that there is little difference between the performances of weighted estimators and split weighted estimators. Nevertheless, the advantage of split weighted estimators is that they can be divided into sub-estimators. This makes optimizing the mixture weights of weighted estimators possible.
\subsection{Naive Mixture Estimators}
The central idea of this paper is to split each estimator into sub-estimators, assign weights to the sub-estimators and optimize the weights. For IS, SWIS, DR and SWDR, the first idea is to split them according to behavior policy. Taking IS for example, it can be rewritten as
\begin{gather}
	\hat{V}_{IS}=\sum_{i=1}^M\frac{n_i}{\sum_{i'=1}^Mn_i'}\hat{V}_{IS,i},
\end{gather}
where $\hat{V}_{IS,i}=\frac{1}{n_i}\sum_{j=1}^{n_i}\sum_{t=0}^T\gamma^t\rho_{i,j,t}r_{i,j,t}$ is estimator for the target value $V$. We then replace $\frac{n_i}{\sum_{i'=1}^Mn_i'}$ with mixture weights $\alpha_i$, form $\hat{V}_{NMIS}=\sum_{i=1}^M\alpha_{i}\hat{V}_{IS,i}$ and optimize the weights. The weights should satisfy $\sum_{i=1}^M\alpha_{i}=1$. The following theorem gives the optimal mixture weights of this problem:

\begin{theorem}
\label{theo:t1}
	Given $M$ unbiased and independent estimators $\hat{V}_1,\hat{V}_2,...,\hat{V}_M$ of a value $V$, the mixture estimator of them with the minimal variance is $\hat{V}_{MIX}=\sum_{i=1}^M\alpha_i^*\hat{V}_i$, where $ \alpha_i^*=\frac{1}{\mathbb{V}[\hat{V}_i]\sum_{i'=1}^M\frac{1}{\mathbb{V}[\hat{V}_{i'}]}}$. The minimal variance is $\frac{1}{\sum_{i=1}^M\frac{1}{\mathbb{V}[\hat{V}_i]}}$.
\end{theorem}

See Appendix \ref{ap:th1} for proof. The condition of independence comes from Assumption \ref{ass:4}. Theorem \ref{theo:t1} is the general case for Section 6 of \citeinline{Aman2017EffectiveEva} as well as the basis of Fixed Effect Model in Meta Analysis \cite{MetaAnalysis}. It can also be applied to SWIS, DR and SWDR. We call this estimator naive mixture estimator because it does not consider the properties of reinforcement learning.
\subsection{Mixture Estimators for Off-Policy Evaluation}
We can further split the estimators by $t$. Taking IS for example, it can be formulated as
\begin{gather}
	\hat{V}_{IS}=\sum_{i=1}^M\sum_{t=0}^T\frac{n_i}{\sum_{i'=1}^Mn_i'}\hat{V}_{IS,i,t},
\end{gather}
where $\hat{V}_{IS,i,t}=\frac{1}{n_i}\sum_{j=1}^{n_i}\gamma^t\rho_{i,j,t}r_{i,j,t}$ is estimator for the value at $t$ denoted by $V_t$. We can replace $\frac{n_i}{\sum_{i'=1}^Mn_i'}$ with $\alpha_{i,t}$ and form $\hat{V}_{MIS}=\sum_{i=1}^M\sum_{t=0}^T\alpha_{i,t}\hat{V}_{IS,i,t}$. The weights should satisfy $\forall t,\ \sum_{i=1}^M\alpha_{i,t}=1$. The following proposition shows how to optimize such mixture weights.
\begin{proposition}
\label{prop:3}
	Denote the covariance matrix of $[\hat{V}_{i,0},\hat{V}_{i,1},...,\hat{V}_{i,T}]$ by $\Sigma_i$. If 
	\begin{itemize}
		\item $\forall i_1,i_2,t_1,t_2$, if $i_1\neq i_2$, then $\hat{V}_{i_1,t_1}$ and $\hat{V}_{i_2,t_2}$ are independent;
		\item $\forall i\forall t$, $\hat{V}_{i,t}$ is unbiased for $V_t$;
		\item $\forall i\ \Sigma_i$ is positive definite;
	\end{itemize}
	then $\hat{V}_{MIXT}=\sum_{i=1}^M\sum_{t=0}^T\alpha_{i,t}\hat{V}_{i,t}$ is unbiased for $V$ and the mixture weights that minimize variance of $\hat{V}_{MIXT}$ are $\overrightarrow{\alpha}_i^*=\Sigma_i^{-1}(\sum_{i'=1}^M\Sigma_{i'}^{-1})^{-1}\overrightarrow{e}$, where $\overrightarrow{\alpha}_i^*=[\alpha_{i,0}^*,\alpha_{i,1}^*,...,\alpha_{i,T}^*]^{T}$ and $\overrightarrow{e}$ is $[1,1,...,1]^T$. Moreover, $\mathbb{V}[\hat{V}_{MIXT}]\le\mathbb{V}[\hat{V}_{MIX}]$ if $\forall i,\ \hat{V}_i=\sum_{t=0}^T\hat{V}_{i,t}$.
\end{proposition}

See appendix \ref{ap:pr3} for proof. The assumption of positive definite is not hard to reach in real world problems. If the reward is constant at some horizon, we can simply remove it from our formula and still get a positive definite covariance matrix. The same results also hold for SWIS, DR and SWDR.
\subsection{$\alpha\beta$ Mixture Estimators for Off-Policy Evaluation}
Compared with IS and SWIS, DR and SWDR both have control variate terms. These terms reduce the variance of estimation \cite{Thomas2016DataEfficientOP}. Taking DR for example, we can divide the estimator to IS plus the control variates and formulate as
\begin{gather}
	\hat{V}_{DR}=\sum_{i=1}^M\sum_{t=0}^T\frac{n_i}{\sum_{i'=1}^Mn_i'}(\hat{V}_{IS,i,t}+\hat{W}_{DR,i,t}),
\end{gather}
where $\hat{W}_{DR,i,t}=\frac{1}{n_i}\sum_{j=1}^{n_i}\gamma^t\left(\rho_{i,j,t-1}\hat{V}(s_{i,j,t})-\rho_{i,j,t}\hat{Q}(s_{i,j,t},a_{i,j,t})\right)$ are estimators for $0$. Similar to the previous sections, we assign $\alpha_{i,t}$ to $\hat{V}_{IS,i,t}$ and $\beta_{i,t}$ to $\hat{W}_{DR,i,t}$ and form $\hat{V}_{\alpha\beta MDR}=\sum_{i=1}^M\sum_{t=0}^T(\alpha_{i,t}\hat{V}_{IS,i,t}+\beta_{i,t}\hat{W}_{DR,i,t})$. The weights should satisfy $\forall t,\ \sum_{i=1}^M\alpha_{i,t}=1$. The following proposition derives optimal mixture weights for mixture estimator with control variates.
\begin{proposition}
\label{prop:5}
	Given $M*(T+1)$ estimators $\hat{V}_{i,t}$ and $M*(T+1)$ estimators $\hat{W}_{i,t}$, if
	\begin{itemize}
		\item  $\forall i_1,i_2,t_1,t_2$, if $i_1\neq i_2$, then $\hat{V}_{i_1,t_1}$ and $\hat{V}_{i_2,t_2}$ are independent, $\hat{W}_{i_1,t_1}$ and $\hat{W}_{i_2,t_2}$ are independent, $\hat{V}_{i_1,t_1}$ and $\hat{W}_{i_2,t_2}$ are independent;
		\item $\forall i,t$, $\mathbb{E}[\hat{V}_{i,t}]=V_t$ and $\mathbb{E}[\hat{W}_{i,t}]=0$;
	\end{itemize}
	then the mixture weights that minimize the variance of mixture estimator with control variates for the estimators are $\overrightarrow{\alpha}_i^*=H_{i,11}(\sum_{i'=1}^MH_{i',11})^{-1}\overrightarrow{e}$, $\overrightarrow{\beta}_i^*=H_{i,21}(\sum_{i'=1}^MH_{i',11})^{-1}\overrightarrow{e}$, where $\left(\begin{matrix}H_{i,11}&H_{i,12}\\H_{i,21}&H_{i,22}\end{matrix}\right)$ is the precision matrix of $[\hat{V}_{i,0}, \hat{V}_{i,1}, ..., \hat{V}_{i,t}, \hat{W}_{i,0}, \hat{W}_{i,1}, ..., \hat{W}_{i,t}]^T$.
\end{proposition}
See Appendix \ref{ap:pr5} for proof. We call this estimator $\alpha\beta$ mixture estimator. The formulations for all three types of mixture estimators can be found in Appendix \ref{ap:ce}.
\section{Variance estimators}
The mixture estimators in Section \ref{sec3} rely on the estimation of variances and covariance matrixes. By Assumption \ref{ass:1} and \ref{ass:2}, we can get strongly consistent variance estimators. Taking $\hat{V}_{IS,i}$ for example, it is formulated as
\begin{gather}
	\hat{V}_{IS,i}=\frac{1}{n_i}\sum_{j=1}^{n_i}\sum_{t=0}^T\gamma^t\rho_{i,j,t}r_{i,j,t}=\frac{1}{n_i}\sum_{j=1}^{n_i}\hat{V}_{IS,i,j},
\end{gather}
which can be interpreted by average of $n_i$ samples of a random variable $\bm{V}_{IS,i}$. So $\mathbb{V}[\hat{V}_{IS,i}]=\frac{1}{n_i}\mathbb{V}[\bm{V}_{IS,i}]$. In this paper, we use half of the data to estimate $\mathbb{V}[\bm{V}_{IS,i}]$, plug the estimated variances into the formulas and estimate the value by the other half of the data. This strategy can be applied to components of IS and DR in all three types of mixture estimators. 

However, for SWIS and SWDR, we can not regard each component as average of samples. Rather, we should regard them as function of average of samples. For example,
\begin{align}
	\hat{V}_{SWDR,i}&=\sum_{j=1}^{n_i}\sum_{t=0}^T\gamma^t\left(u_{i,j,t-1}\hat{V}(s_{i,j,t})+u_{i,j,t}(r_{i,j,t}-\hat{Q}(s_{i,j,t},a_{i,j,t}))\right)\notag\\
	&=\sum_{t=0}^T\gamma^t\left(\frac{\sum_{j=1}^{n_i}\rho_{i,j,t-1}\hat{V}(s_{i,j,t})}{\sum_{j=1}^{n_i}\rho_{i,j,t-1}}+\frac{\sum_{j=1}^{n_i}\rho_{i,j,t}(r_{i,j,t}-\hat{Q}(s_{i,j,t},a_{i,j,t}))}{\sum_{j=1}^{n_i}\rho_{i,j,t}}\right)\notag\\
	&=\sum_{t=0}^T\gamma^t\left(\frac{\frac{1}{n_i}\sum_{j=1}^{n_i}\rho_{i,j,t-1}\hat{V}(s_{i,j,t})}{\frac{1}{n_i}\sum_{j=1}^{n_i}\rho_{i,j,t-1}}+\frac{\frac{1}{n_i}\sum_{j=1}^{n_i}\rho_{i,j,t}(r_{i,j,t}-\hat{Q}(s_{i,j,t},a_{i,j,t}))}{\frac{1}{n_i}\sum_{j=1}^{n_i}\rho_{i,j,t}}\right)\notag\\
	&\triangleq \sum_{t=0}^T\left(\frac{\frac{1}{n_i}\sum_{j=1}^{n_i}\hat{X}_{i,j,t}}{\frac{1}{n_i}\sum_{j=1}^{n_i}\hat{W}_{i,j,t}}+\frac{\frac{1}{n_i}\sum_{j=1}^{n_i}\hat{Z}_{i,j,t}}{\frac{1}{n_i}\sum_{j=1}^{n_i}\hat{Y}_{i,j,t}}\right)
\end{align}
We approximate $\mathbb{V}[\hat{V}_{SWDR,i,t}]$ by Delta Method \cite{Owen2013MonteCarlo}. See appendix \ref{asdm} for introduction. 

The variance and covariance estimators for components of all the estimators are in Appendix \ref{ap:ce3}.

\section{Related Work}
	Recent advances in OPE can be split into two categories: Importance Sampling based OPE and stationary distribution based OPE. For Importance Sampling based OPE, previous works about mixture estimators mainly focus on mixing different kinds of estimators. \citeinline{Thomas2016DataEfficientOP} optimize a mixture weight between WDR and Direct Method to reduce the MSE of estimation. Following it, recent works \cite{WangAD17,DBLP:journals/corr/abs-1907-09623,Su2019CABCA,Su2020AdaptiveES} propose more strategies to blend different off-policy evaluation algorithms. The mixture estimators in this paper are different from theirs. They mix different kinds of estimators, while we mix estimators of different behavior policies. \citeinline{Aman2017EffectiveEva} derive Weighted IPS estimators, which is actually NMIS estimators for contextual bandits. Compared with them, we not only generalize the idea to finite-horizon MDP, but also apply our techniques to more OPE algorithms. Additionally, Balanced IPS estimator \cite{Aman2017EffectiveEva} or Multiple Importance Sampling \cite{Owen2013MonteCarlo} is another way to reduce variance for OPE of contextual bandits with multiple behavior policies. However, as far as we know, no previous work has generalized it to finite-horizon OPE so we do not compare with it. Stationary distribution based OPE are built on the estimation of stationary distributions or ratio of stationary distributions \cite{Liu2018BreakingTC, Xie2019OptimalOE,Tang2019DoublyRB,DBLP:journals/corr/abs-1910-12809, DBLP:conf/iclr/Mousavi0LZ20,DBLP:conf/iclr/ZhangD0S20}. To evaluate with multiple behavior policies, \citeinline{Nachum2019DualDICEBE} and \citeinline{DChenWHGZ20} build the mixture policy of the behavior policies and estimate the ratio of stationary distributions. Different from them, we optimize the mixture weights of estimators while they regard data from different policies equally. It would be interesting to see how our methods help improve these methods.

\section{Experiments}
\subsection{Experimental Settings}
We construct a simulated recommender platform based on RecSim \cite{Ie2019RecSimAC}. The interactions in this platform form a Partially Observable Markov Decision Process(POMDP) \cite{Sutton1998ReinforcementLA}. Detailed configuration of this environment can be found in Appendix \ref{ap:es}. We implemented IS, WIS, SWIS, DR, WDR and SWDR as baselines. We call naive mixture estimators for the methods NMIS, NMWIS, NMDR and NMWDR. Mixture estimators for the methods are called MIS, MWIS, MDR, MWDR. $\alpha\beta$ mixture estimators are called $\alpha\beta$ MDR and $\alpha\beta$ MWDR. See Appendix \ref{ap:id} for implementation of the OPE algorithms. For mixture estimators and $\alpha\beta$ mixture estimators, we choose a hyper-parameter $T$, mix the values from 0 to T and simply add up the remains. See Appendix \ref{ap:ih} for discussion about it.

\begin{figure}
\begin{minipage}{0.25\textwidth}
  \centering
  \includegraphics[width=.95\linewidth]{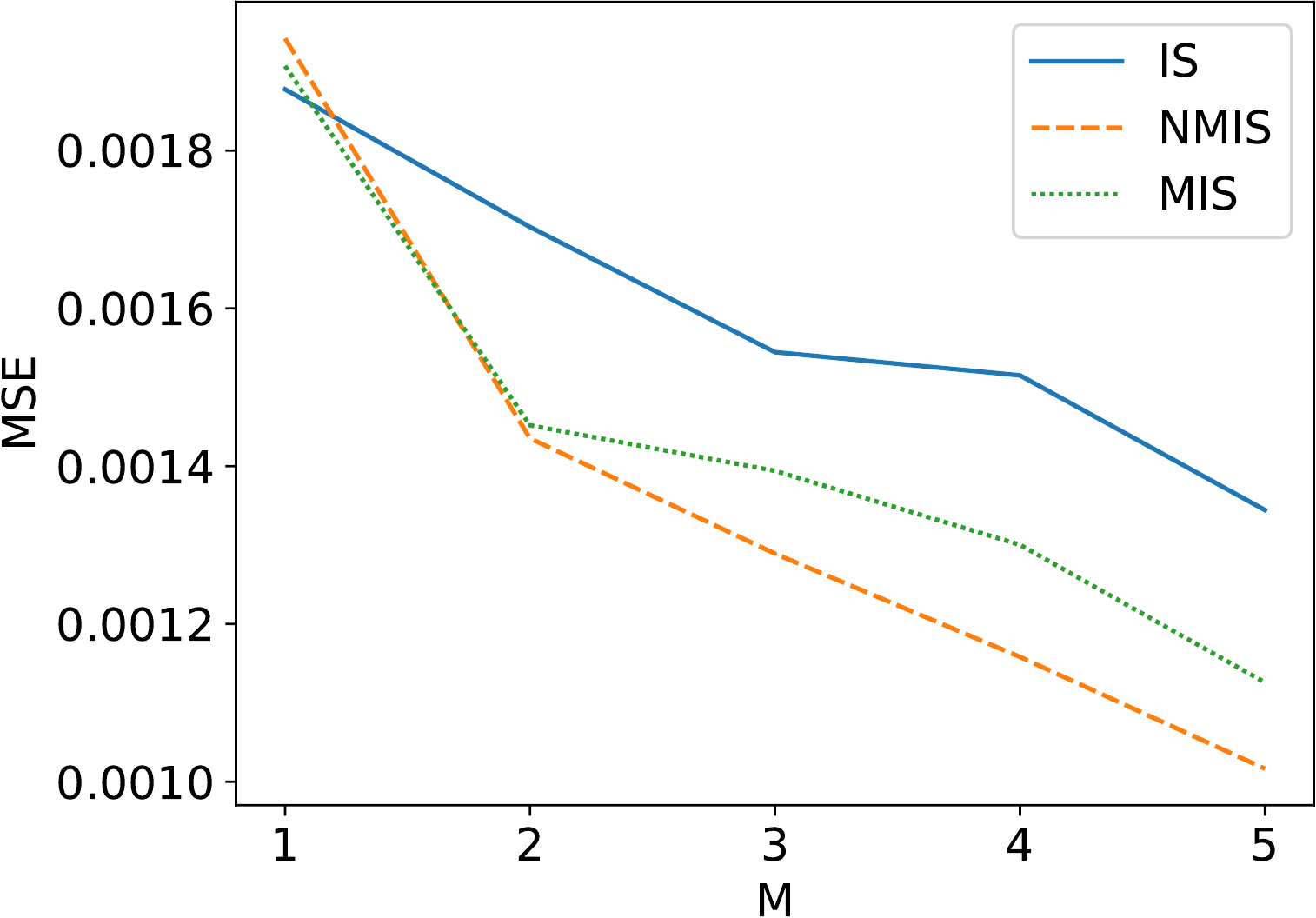}
\end{minipage}\hfill
\begin{minipage}{0.25\textwidth}
  \centering
  \includegraphics[width=.95\linewidth]{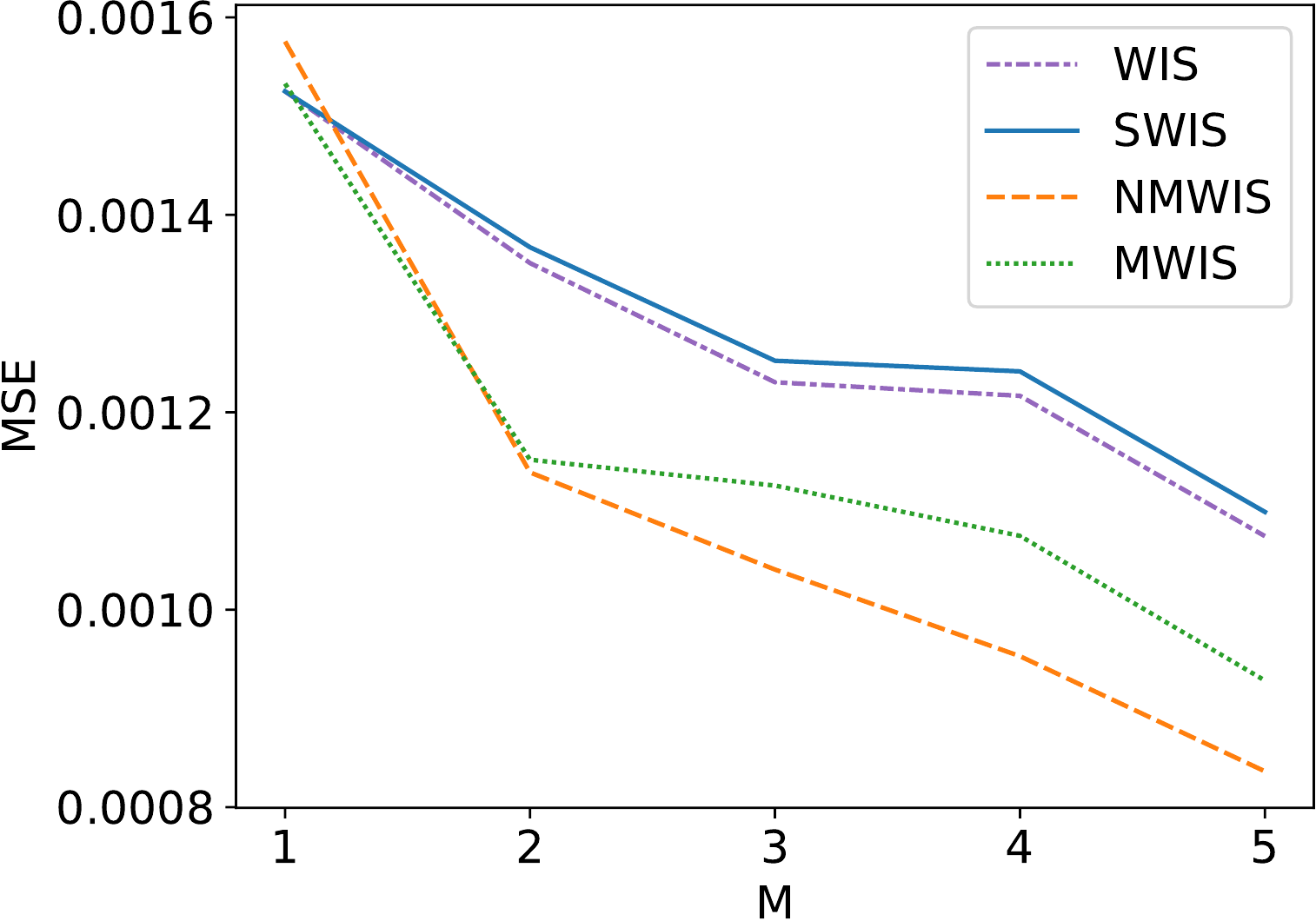}
\end{minipage}\hfill
\begin{minipage}{0.25\textwidth}
  \centering
  \includegraphics[width=.95\linewidth]{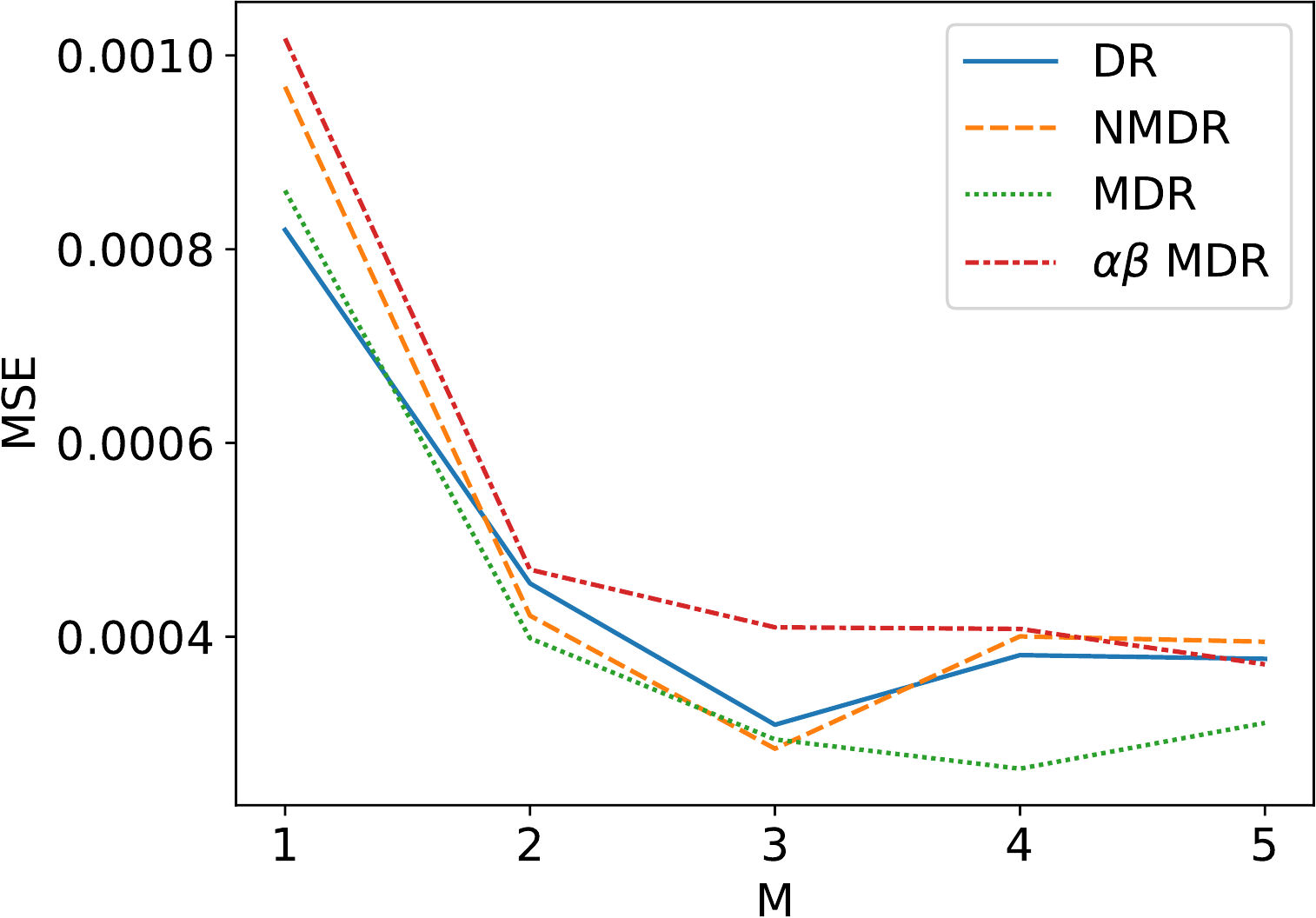}
\end{minipage}\hfill
\begin{minipage}{0.25\textwidth}
  \centering
  \includegraphics[width=.95\linewidth]{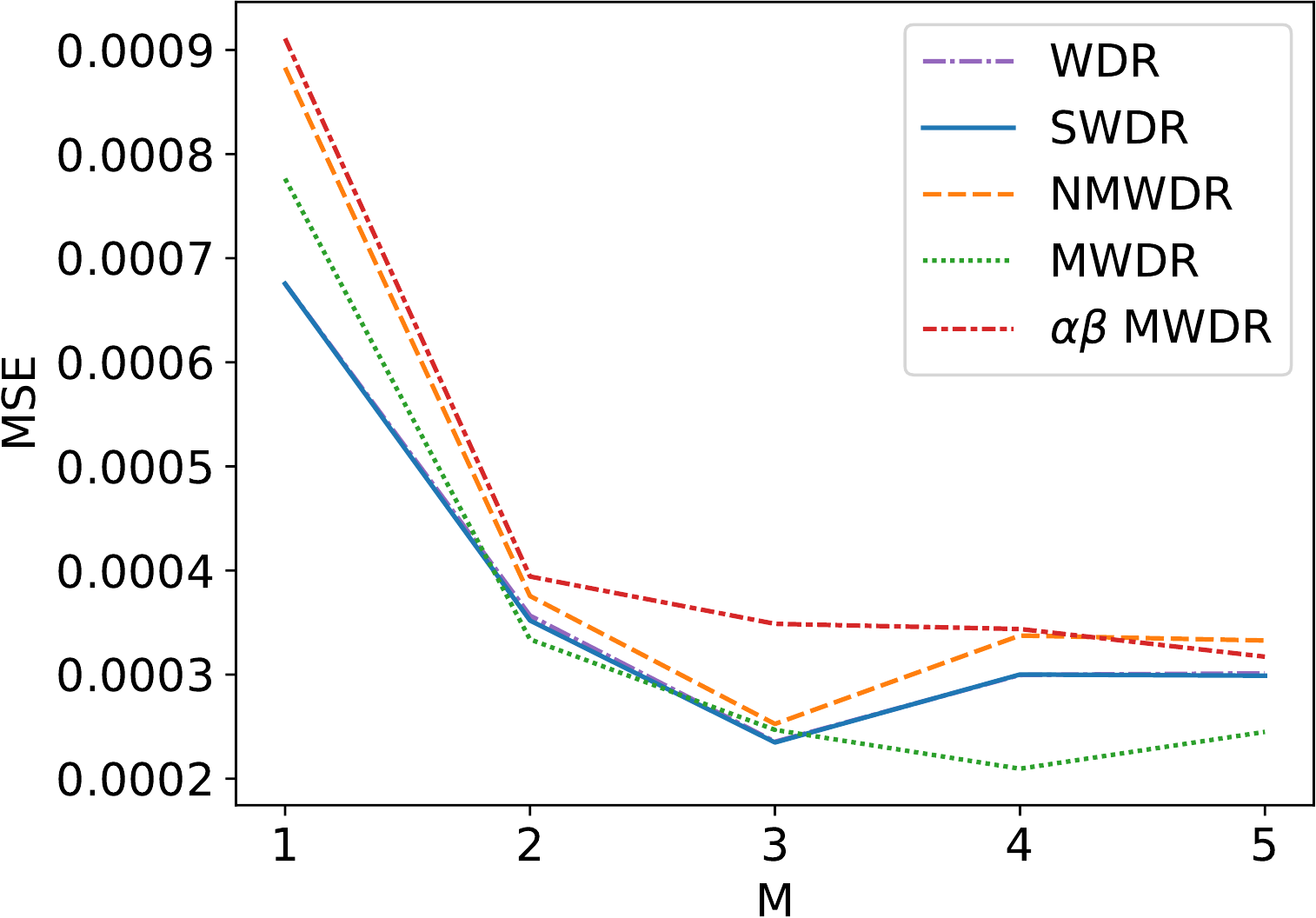}
\end{minipage}

  \caption{MSE of all four types of estimators.}
  \label{fig:ave}
\end{figure}
\subsection{Results}
 With the chosen T, the MSE of the estimators with different M on test set are plotted in Figure \ref{fig:ave}. Numerical results can be found in Appendix \ref{ap:am}. In each figure, when M=1, the baselines have the lowest MSE. This is because only half of the samples in mixture estimators are used to estimate values. As M increases, the MSE of all estimators decrease. For IS and WIS, when M=5, both naive mixture estimators and mixture estimators are better than baselines. However, naive mixture estimators have the best results. There are two possible reasons for it. First, mixture estimators only mix the first several values while naive mixture estimators mix values of the whole horizon. Second, mixture estimators require estimation of covariance matrix, which may amplify the error of estimation. For DR and WDR, when M=5, mixture estimators produce the best results while naive mixture estimators and $\alpha\beta$ mixture estimators produce comparable results with baselines. This indicates that $\alpha\beta$ mixture estimators are not as effective as theory. We compute the average condition number for the estimated covariance matrixes of all the mixture estimators. See Appendix \ref{ap:mcd}. With relatively large condition numbers, the error of $\alpha\beta$ mixture estimators is amplified in matrix inversion.

\section{Conclusion}
We derive naive mixture estimators, mixture estimators and $\alpha\beta$ mixture estimators for OPE with multiple behavior policies. To estimate the mixture weights for weighted estimators, we introduce Delta Method to estimate the variances and covariances of weighted estimators. In our experiments on simulated recommender systems, we show that naive mixture estimators and mixture estimators are effective in reducing MSE while $\alpha\beta$ mixture estimators suffer from ill covariance matrixes. Future work can focus on mitigating this problem.

\bibliography{reference}
\newpage

\appendix
\section{Proof of Theorems and Propositions}
\subsection{Proof of Theorem 1}
\label{ap:th1}
	The variance of a mixture estimator is
	\begin{align*}
		\mathbb{V}[\hat{V}_{\alpha}]=\sum_{i=1}^M\alpha_i^2\mathbb{V}[\hat{V}_i].
	\end{align*}
	The variance of $\hat{V}_{MIX}$ is
	\begin{align}
		\mathbb{V}[\hat{V}_{MIX}]&=\sum_{i=1}^M\frac{1}{(\mathbb{V}[\hat{V}_i]\sum_{j=1}^M\frac{1}{\mathbb{V}[\hat{V}_j]})^2}\mathbb{V}[\hat{V}_i]\notag\\
		&=\frac{1}{\sum_{i=1}^M\frac{1}{\mathbb{V}[\hat{V}_i]}}.
	\end{align}
	Therefore, 
	\begin{align}
		\frac{\mathbb{V}[\hat{V}_{\alpha}]}{\mathbb{V}[V_{MIX}]}&=\sum_{i=1}^M\frac{1}{\mathbb{V}[\hat{V}_i]}\sum_{i=1}^M\alpha_i^2\mathbb{V}[\hat{V}_i]\notag\\
		&\ge(\sum_{i=1}^M|\alpha_i|)^2\notag\\
		&\ge(|\sum_{i=1}^M\alpha_i|)^2=1.
	\end{align}
	This means that any mixture estimator other than $\hat{V}_{MIX}$ has higher or equal variance.
\subsection{Proof of Proposition 1}
\label{ap:pr3}
	The unbiasedness is proved as
	\begin{align}
		\mathbb{E}[\hat{V}_{MIXT}-V]&=\mathbb{E}[\sum_{t=0}^{T}\sum_{i=1}^M\alpha_{i,t}\hat{V}_{i,t}-\sum_{t=0}^{T}V_t]\notag\\
		&=\mathbb{E}[\sum_{t=0}^{T}\sum_{i=1}^M\alpha_{i,t}(\hat{V}_{i,t}-V_t)]\notag\\
		&=\sum_{t=0}^{T}\sum_{i=1}^M\alpha_{i,t}\mathbb{E}[(\hat{V}_{i,t}-V_t)]\notag\\
		&=0.
	\end{align}
	The variance of $\hat{V}_{MIXT}$ is
	\begin{align}
	\label{va:1}
		\mathbb{V}[\hat{V}_{MIXT}]=\sum_{i=1}^M\left(\sum_{t=0}^T\alpha_{i,t}^2\mathbb{V}[\hat{V}_{i,t}]+2\sum_{1\le t_1<t_2\le T}\alpha_{i,t_1}\alpha_{i,t_2}Cov[\hat{V}_{i,t_1},\hat{V}_{i,t_2}]\right),
	\end{align}
	where $\forall t$, $\sum_{i=1}^M\alpha_{i,t}=1$. We construct the Lagrangian function for the problem
	\begin{align}
		\mathcal{L}[A,\Lambda]=\sum_{i=1}^M\left(\sum_{t=0}^T\alpha_{i,t}^2\mathbb{V}[\hat{V}_{i,t}]+2\sum_{1\le t_1<t_2\le T}\alpha_{i,t_1}\alpha_{i,t_2}Cov[\hat{V}_{i,t_1},\hat{V}_{i,t_2}]\right)-\sum_{t=0}^T\lambda_t(\sum_{i=1}^M\alpha_{i,t}-1),
	\end{align}
	where $\lambda_t$ are Lagrangian multipliers. Let $\frac{\partial \mathcal{L}}{\partial \alpha_{i,t}}=0$ and $\frac{\partial \mathcal{L}}{\partial \lambda_t}=0$, we get
	\begin{gather}
	\label{la:1}
		\forall i\forall t\quad 2\mathbb{V}[\hat{V}_{i,t}]\alpha_{i,t}+2\sum_{\tau\neq t}\alpha_{i,\tau}Cov[\hat{V}_{i,t},\hat{V}_{i,\tau}]=\lambda_t,\\
		\forall t\quad \sum_{i=1}^M\alpha_{i,t}=1.
		\label{la:2}
	\end{gather}
	We denote $[\lambda_0,\lambda_1,...,\lambda_T]^T$ by $\overrightarrow{\Lambda}$. From (\ref{la:1}) we know that $\forall i\ 2\Sigma_i\overrightarrow{\alpha}_i^*=\overrightarrow{\Lambda}$, which means $\forall i\ \overrightarrow{\alpha}_i^*=\frac{1}{2}\Sigma_i^{-1}\overrightarrow{\Lambda}$. Add up (\ref{la:2}), and we get $\overrightarrow{e}=\sum_{i=1}^M\overrightarrow{\alpha}_i^*=\frac{1}{2}\sum_{i=1}^M\Sigma_i^{-1}\overrightarrow{\Lambda}$, which means $\frac{1}{2}\overrightarrow{\Lambda}=(\sum_{i=1}^M\Sigma_i^{-1})^{-1}\overrightarrow{e}$. Therefore, $\forall i\ \overrightarrow{\alpha}_i^*=\Sigma_i^{-1}(\sum_{i=1}^M\Sigma_i^{-1})^{-1}\overrightarrow{e}$. We have found a stationary point of (\ref{va:1}), so now we need to show that it is the global minima. Note that the solutions of (\ref{la:2}) forms a convex set on $R^{M\times T}$. In addition, because all the covariance matrixes are positive definite, $\mathbb{V}[\hat{V}_{MIXT}]$ is also a strictly convex function of $\overrightarrow{\alpha}$. Therefore, $\overrightarrow{\alpha}_i^*$ are the local minima as well as the global minima. 
	
To prove the variance reduction, we rewrite $\hat{V}_{MIX}$ as
	\begin{gather}
		\hat{V}_{MIX} = \sum_{i=1}^M\sum_{t=0}^{T}\alpha_i V_{i,t}.
	\end{gather}
	Note that $\alpha_{i,t}$ are the optimal mixture weights to minimize variance, so $\mathbb{V}[\hat{V}_{MIXT}]\le\mathbb{V}[\hat{V}_{MIX}]$.
\subsection{Proof of Proposition 2}
\label{ap:pr5}
The variance of $V_{MIXC}$ is 
\begin{align}
	\mathbb{V}[\hat{V}_{MIXC}]&=\sum_{i=1}^M\sum_{t_1=0}^T\sum_{t_2=0}^T(\alpha_{i,t_1}\alpha_{i,t_2}Cov[\hat{V}_{i,t_1},\hat{V}_{i,t_2}]+\beta_{i,t_1}\beta_{i,t_2}Cov[\hat{W}_{i,t_1},\hat{W}_{i,t_2}]\notag\\
	&\quad +\alpha_{i,t_1}\beta_{i,t_2}Cov[\hat{V}_{i,t_1},\hat{W}_{i,t_2}]+\beta_{i,t_1}\alpha_{i,t_2}Cov[\hat{W}_{i,t_1},\hat{V}_{i,t_2}]).
\end{align}
Like in \ref{ap:pr3}. We construct the Lagrangian function for it by
\begin{align}
	\mathcal{L}[A,B,\Lambda]&=\sum_{i=1}^M\sum_{t_1=0}^T\sum_{t_2=0}^T(\alpha_{i,t_1}\alpha_{i,t_2}Cov[\hat{V}_{i,t_1},\hat{V}_{i,t_2}]+\beta_{i,t_1}\beta_{i,t_2}Cov[\hat{W}_{i,t_1},\hat{W}_{i,t_2}]\notag\\
	&\quad +\alpha_{i,t_1}\beta_{i,t_2}Cov[\hat{V}_{i,t_1},\hat{W}_{i,t_2}]+\beta_{i,t_1}\alpha_{i,t_2}Cov[\hat{W}_{i,t_1},\hat{V}_{i,t_2}])\notag\\
	&\quad -\sum_{t=0}^T\lambda_t(\sum_{i=1}^M\alpha_{i,t}-1).
\end{align}
Let $\frac{\partial \mathcal{L}}{\partial  \overrightarrow{\alpha}}=0$, $\frac{\partial \mathcal{L}}{\partial  \overrightarrow{\beta}}=0$, $\frac{\partial \mathcal{L}}{\partial  \overrightarrow{\Lambda}}=0$, we have
\begin{gather}
	\label{formula:f328}
	\forall i,\ 2\left(\begin{matrix}\overrightarrow{\alpha}_i^*\\ \overrightarrow{\beta}_i^*\end{matrix}\right) =\left(\begin{matrix}H_{i,11}&H_{i,12}\\H_{i,21}&H_{i,22}\end{matrix}\right)\left(\begin{matrix}\overrightarrow{\Lambda}\\ \overrightarrow{0}\end{matrix}\right),\\
	\forall t\quad \sum_{i=1}^M\alpha_{i,t}=1.
	\label{formula:f329}
\end{gather}
By (\ref{formula:f329}) we can get $\overrightarrow{e}=\sum_{i=1}^M\overrightarrow{\alpha}_i^*=\frac{1}{2}\sum_{i=1}^MH_{i,11}\overrightarrow{\Lambda}$ and compute $\overrightarrow{\Lambda}$. Bringing it to (\ref{formula:f328}), we get $\overrightarrow{\alpha}_i^*=H_{i,11}(\sum_{j=1}^MH_{j,11})^{-1}\overrightarrow{e}$ and $\overrightarrow{\beta}_i^*=H_{i,21}(\sum_{j=1}^MH_{j,11})^{-1}\overrightarrow{e}$. Similar to \ref{ap:pr3}, we can prove that this stationary point is the global minima.
\section{Delta Method}
\label{asdm}
\subsection{Derivation of Delta Method}
In this section, we introduce Delta Method \cite{Owen2013MonteCarlo}, which is used for estimating the variance of WIS and WDR. Given a function of expectations of random variables $\theta=f(\mathbb{E}[\bm{X}])$, it is usually hard to obtain an unbiased estimator. Luckily, if $f$ is a continuous function, the empirical estimation $\hat{\theta}=f(\overline{\bm{X}})$, where $\overline{\bm{X}}$ is the sample mean of data, is strongly consistent. However, the samples are inside the function so the variance of $\hat{\theta}$ is hard to determine. We present the technology in \citeinline{Owen2013MonteCarlo} here to build the framework of estimating variance for $\hat{\theta}$.

By first order Taylor expansion of $f$, $\hat{\theta}$ is approximated by 
\begin{align}
	\hat{\theta}\approx \theta+\sum_{i=1}^d(\overline{X}_i-\mu_i)f_i(\mu),
\end{align} 
where $X_i$ is the $i$-th component of $\bm{X}$, $\mu_i=\mathbb{E}[X_i]$, and $f_i=\frac{\partial f}{\partial X_i}$. When the sample size n is large, $\overline{X}_i$ is very close to $\mu_i$, so the right hand side is a good approximate of $\hat{\theta}$. The variance of $\hat{\theta}$ is then approximated by 
\begin{align}
\label{ap:delta_method}
\mathbb{V}[\hat{\theta}]\approx \frac{1}{n}\left(\sum_{i=1}^df_i(\mu)^2\sigma_i^2+2\sum_{i=1}^{d-1}\sum_{j=i+1}^df_i(\mu)f_j(\mu)\sigma_{i,j}\right)
	=\frac{1}{n}(\nabla f)^T\Sigma(\nabla f),	
\end{align} 
where $\sigma_i^2$ is the variance of $X_i$, $\sigma_{i,j}$ is the covariance of $X_i$ and $X_j$, $\nabla f$ is the gradient of $f$ and $\Sigma$ is the corresponding covariance matrix. 
\subsection{Variance of Weighted Estimators}
\label{ap:vw}
We estimate the variance of weighted estimators by (\ref{ap:delta_method}). Define $\theta=f(\mu_x,\mu_y)=\frac{\mu_y}{\mu_x}$, where $\mu_x=\mathbb{E}[\bm{X}]$ and $\mu_y=\mathbb{E}[\bm{Y}]$, then $f_{x}=-\frac{\mu_y}{\mu_x^2}$, $f_{y}=\frac{1}{\mu_x}$. If we define $\sigma_x^2=\mathbb{V}[\bm{X}]$, $\sigma_y^2=\mathbb{V}[\bm{Y}]$, and $\sigma_{x,y}=Cov[\bm{X},\bm{Y}]$, $\mathbb{V}[\hat{\theta}]$ is approximated by
\begin{align}
	\mathbb{V}[\hat{\theta}]&=\frac{1}{n}\left(\frac{\mu_y^2\sigma_x^2}{\mu_x^4}+\frac{\sigma_y^2}{\mu_x^2}-\frac{2\mu_y\sigma_{x,y}}{\mu_x^3}\right)\notag\\
	&=\frac{1}{n}\frac{\theta^2\sigma_x^2+\sigma_y^2-2\theta\sigma_{x,y}}{\mu_x^2}\notag\\
	&=\frac{1}{n}\frac{\theta^2\sigma_x^2+\sigma_y^2-2\theta\sigma_{x,y}+(\mu_y-\theta \mu_x)^2}{\mu_x^2}\notag\\
	&=\frac{1}{n}\frac{\mathbb{E}[\theta^2\bm{X}^2]+\mathbb{E}[\bm{Y}^2]-2\mathbb{E}[\theta\bm{X}\bm{Y}]}{\mu_x^2}\notag\\
	&=\frac{1}{n}\frac{\mathbb{E}[(\theta\bm{X}-\bm{Y})^2]}{\mu_x^2}.
\end{align}
\subsection{Covariance of Weighted Estimators}
\label{ap:cw}
To estimate the covariance, we define $\theta_1=f(\mu_{x_1},\mu_{y_1})=\frac{\mu_{y_1}}{\mu_{x_1}}$ and $\theta_2=f(\mu_{x_2},\mu_{y_2})=\frac{\mu_{y_2}}{\mu_{x_2}}$. The corresponding expectations, variances and covariances are represented by $\mu_{x_1}$, $\mu_{y_1}$, $\mu_{x_2}$, $\mu_{y_2}$, $\sigma_{x_1}^2$, $\sigma_{y_1}^2$, $\sigma_{x_2}^2$, $\sigma_{y_2}^2$, $\sigma_{x_1,x_2}$, $\sigma_{x_1,y_2}$, $\sigma_{y_1,x_2}$, $\sigma_{y_1,y_2}$. The covariance approximation $Cov[\hat{\theta_1},\hat{\theta_2}]$ is 
\begin{align}
	Cov[\hat{\theta}_1,\hat{\theta}_2]&=\frac{1}{n}\left(\frac{\mu_{y_1}\mu_{y_2}\sigma_{x_1,x_2}}{\mu_{x_1}^2\mu_{x_2}^2}-\frac{\mu_{y_1}\sigma_{x_1,y_2}}{\mu_{x_1}^2\mu_{x_2}}-\frac{\mu_{y_2}\sigma_{y_1,x_2}}{\mu_{x_1}\mu_{x_2}^2}+\frac{\sigma_{y_1,y_2}}{\mu_{x_1}\mu_{x_2}}\right)\notag\\
	&=\frac{1}{n}\frac{\theta_1\theta_2\sigma_{x_1,x_2}-\theta_1\sigma_{x_1,y_2}-\theta_2\sigma_{y_1,x_2}+\sigma_{y_1,y_2}}{\mu_{x_1}\mu_{x_2}}\notag\\
	&=\frac{1}{n}\frac{\theta_1\theta_2\sigma_{x_1,x_2}-\theta_1\sigma_{x_1,y_2}-\theta_2\sigma_{y_1,x_2}+\sigma_{y_1,y_2}+(\mu_{y_1}-\theta_1\mu_{x_1})(\mu_{y_2}-\theta_2\mu_{x_2})}{\mu_{x_1}\mu_{x_2}}\notag\\
	&=\frac{1}{n}\frac{\mathbb{E}[\theta_1\theta_2\bm{X}_1\bm{X}_2]-\mathbb{E}[\theta_1\bm{X}_1\bm{Y}_2]-\mathbb{E}[\theta_2\bm{Y}_1\bm{X}_2]+\mathbb{E}[\bm{Y_1}\bm{Y}_2]}{\mu_{x_1}\mu_{x_2}}\notag\\
	&=\frac{1}{n}\frac{\mathbb{E}[(\theta_1\bm{X}_1-\bm{Y}_1)(\theta_2\bm{X}_2-\bm{Y}_2)]}{\mu_{x_1}\mu_{x_2}}.
\end{align}
\subsection{Variance of Summation of Weighted Estimators}
\label{ap:vsw}
We use the formulas in \ref{ap:vw} and \ref{ap:cw} to derive the variance of summation of weighted estimators. 
Define $\theta=g(\mu_{x_1},\mu_{y_1},\mu_{x_2},\mu_{y_2},...,\mu_{x_T},\mu_{y_T})=\sum_{t=0}^T\theta_{t}=\sum_{t=0}^T\frac{\mu_{y_t}}{\mu_{x_t}}$, where $\mu_{x_t}=\mathbb{E}[\bm{X}_t]$ and $\mu_{y_t}=\mathbb{E}[\bm{Y}_t]$. Then
\begin{align}
	\mathbb{V}[\hat{\theta}]&\approx\sum_{t=0}^T\mathbb{V}[\hat{\theta}_t]+2\sum_{t=0}^{T-1}\sum_{\tau=t+1}^T Cov[\hat{\theta}_t,\hat{\theta}_{\tau}]\notag\\
	&=\frac{1}{n}\left(\sum_{t=0}^T\frac{\mathbb{E}[(\theta_t\bm{X}_t-\bm{Y}_t)^2]}{\mu_{x_t}^2}+2\sum_{t=0}^{T-1}\sum_{\tau=t+1}^T\frac{\mathbb{E}[(\theta_t\bm{X}_t-\bm{Y}_t)(\theta_{\tau}\bm{Y}_{\tau}-\bm{X}_{\tau})]}{\mu_{x_t}\mu_{x_{\tau}}}\right)\notag\\
	&=\frac{1}{n}\mathbb{E}\left[(\sum_{t=0}^T\frac{\theta_t\bm{X}_t-\bm{Y}_t}{\mu_{x_t}})^2\right].
\end{align}
If we add the subscript $i$ to the formula, the formula would be
\begin{align}
	\mathbb{V}[\hat{V}_i]&\approx\frac{1}{n_i}\mathbb{E}\left[(\sum_{t=0}^T\frac{\theta_{i,t}\bm{X}_{i,t}-\bm{Y}_{i,t}}{\mu_{x_{i,t}}})^2\right].
\end{align}
If we use the samples $X_{i,j,t}$ and $Y_{i,j,t}$ to estimate $\mathbb{V}[\hat{V}_i]$, the estimated variance would be
\begin{align}
	Var_i&=\frac{1}{n_i^2}\sum_{j=1}^{n_i}\left(\sum_{t=0}^T\frac{\hat{Y}_{i,j,t}-\hat{\theta}_{i,t}\hat{X}_{i,j,t}}{\frac{1}{n_i}\sum_{k=1}^{n_i}\hat{X}_{i,k,t}}\right)^2\notag\\
	&=\sum_{j=1}^{n_i}\left(\sum_{t=0}^T\frac{\hat{Y}_{i,j,t}-\hat{\theta}_{i,t}\hat{X}_{i,j,t}}{\sum_{k=1}^{n_i}\hat{X}_{i,k,t}}\right)^2.
\end{align}
We now show that $n_i*Var_i$ is strongly consistent. It is derived as
\begin{align}
	n_i*Var_i&=\frac{1}{n_i}\sum_{j=1}^{n_i}\left(\sum_{t=0}^T\frac{\hat{Y}_{i,j,t}-\hat{\theta}_{i,t}\hat{X}_{i,j,t}}{\frac{1}{n_i}\sum_{k=1}^{n_i}\hat{X}_{i,k,t}}\right)^2\notag\\
	&=\frac{1}{n_i}\sum_{j=1}^{n_i}\sum_{t=0}^T\sum_{\tau=0}^T\frac{\hat{Y}_{i,j,t}-\hat{\theta}_{i,t}\hat{X}_{i,j,t}}{\frac{1}{n_i}\sum_{k=1}^{n_i}\hat{X}_{i,k,t}}\cdot \frac{\hat{Y}_{i,j,\tau}-\hat{\theta}_{i,\tau}\hat{X}_{i,j,\tau}}{\frac{1}{n_i}\sum_{k=1}^{n_i}\hat{X}_{i,k,\tau}}\notag\\
	&=\sum_{t=0}^T\sum_{\tau=0}^T\frac{\frac{1}{n_i}\sum_{j=1}^{n_i}(\hat{Y}_{i,j,t}-\hat{\theta}_{i,t}\hat{X}_{i,j,t})(\hat{Y}_{i,j,\tau}-\hat{\theta}_{i,\tau}\hat{X}_{i,j,\tau})}{\frac{1}{n_i}\sum_{k=1}^{n_i}\hat{X}_{i,k,t}\cdot \frac{1}{n_i}\sum_{k=1}^{n_i}\hat{X}_{i,k,\tau}}\notag\\
	&=\sum_{t=0}^T\sum_{\tau=0}^T\frac{\frac{1}{n_i}\sum_{j=1}^{n_i}\hat{Y}_{i,j,t}\hat{Y}_{i,j,\tau}+\frac{\hat{\theta}_{i,t}\hat{\theta}_{i,\tau}}{n_i}\sum_{j=1}^{n_i}\hat{X}_{i,j,t}\hat{X}_{i,j,\tau}}{\frac{1}{n_i}\sum_{k=1}^{n_i}\hat{X}_{i,k,t}\cdot \frac{1}{n_i}\sum_{k=1}^{n_i}\hat{X}_{i,k,\tau}}\notag\\
	&\quad -\frac{\frac{\hat{\theta}_{i,t}}{n_i}\sum_{j=1}^{n_i}\hat{X}_{i,j,t}\hat{Y}_{i,j,\tau}+\frac{\hat{\theta}_{i,\tau}}{n_i}\sum_{j=1}^{n_i}\hat{X}_{i,j,\tau}\hat{Y}_{i,j,t}}{\frac{1}{n_i}\sum_{k=1}^{n_i}\hat{X}_{i,k,t}\cdot \frac{1}{n_i}\sum_{k=1}^{n_i}\hat{X}_{i,k,\tau}}.
\end{align}
Therefore, $n_i*Var_i$ is strongly consistent for
\begin{align}
	E_{i}&=\sum_{t=0}^T\sum_{\tau=0}^T\frac{\mathbb{E}[\bm{Y}_{i,t}\bm{Y}_{i,\tau}]+\theta_{i,t}\theta_{i,\tau}\mathbb{E}[\bm{X}_{i,t}\bm{X}_{i,\tau}]-\theta_{i,t}\mathbb{E}[\bm{X}_{i,t}\bm{Y}_{i,\tau}]-\theta_{i,\tau}\mathbb{E}[\bm{X}_{i,\tau}\bm{Y}_{i,t}]}{\mu_{x_{i,t}}\mu_{x_{i,\tau}}}\notag\\
	&=\sum_{t=0}^T\sum_{\tau=0}^T\frac{\mathbb{E}[(\bm{Y}_{i,t}-\theta_{i,t}\bm{X}_{i,t})(\bm{Y}_{i,\tau}-\theta_{i,\tau}\bm{X}_{i,\tau})]}{\mu_{x_{i,t}}\mu_{x_{i,\tau}}}\notag\\
	&=\mathbb{E}\left[(\sum_{t=0}^T\frac{\bm{Y}_{i,t}-\theta_{i,t}\bm{X}_{i,t}}{\mu_{x_{i,t}}})^2\right].
\end{align}
\subsection{Covariance of Summation of two Weighted Estimators}
\label{ap:csw}

Similarly, we can estimate the covariance of $\hat{\theta}_1$ and $\hat{\theta}_2$, where $\hat{\theta}_1$ estimates $\theta_1=g(\mu_{w_1},\mu_{x_1},\mu_{y_1},\mu_{z_1})=\frac{\mu_{x_1}}{\mu_{w_1}}+\frac{\mu_{z_1}}{\mu_{y_1}}$ and $\hat{\theta}_2$ estimates $\theta_2=g(\mu_{w_2},\mu_{x_2},\mu_{y_2},\mu_{z_2})=\frac{\mu_{x_2}}{\mu_{w_2}}+\frac{\mu_{z_2}}{\mu_{y_2}}$. Define $\theta_{11}=\frac{\mu_{x_1}}{\mu_{w_1}}$, $\theta_{12}=\frac{\mu_{z_1}}{\mu_{y_1}}$, $\theta_{21}=\frac{\mu_{x_2}}{\mu_{w_2}}$, $\theta_{22}=\frac{\mu_{z_2}}{\mu_{y_2}}$, then
\begin{align}
	Cov[\hat{\theta}_1,\hat{\theta}_2]&=Cov[\hat{\theta}_{11},\hat{\theta}_{21}]+Cov[\hat{\theta}_{11},\hat{\theta}_{22}]+Cov[\hat{\theta}_{12},\hat{\theta}_{21}]+Cov[\hat{\theta}_{12},\hat{\theta}_{22}]\notag\\
	&=\frac{1}{n}(\frac{\mathbb{E}[(\theta_{11}\bm{W}_1-\bm{X}_1)(\theta_{21}\bm{W}_2-\bm{X}_2)]}{\mu_{w_1}\mu_{w_2}}+\frac{\mathbb{E}[(\theta_{11}\bm{W}_1-\bm{X}_1)(\theta_{22}\bm{Y}_2-\bm{Z}_2)]}{\mu_{w_1}\mu_{y_2}}+\notag\\
	&\quad\quad\quad \frac{\mathbb{E}[(\theta_{12}\bm{Y}_1-\bm{Z}_1)(\theta_{21}\bm{W}_2-\bm{X}_2)]}{\mu_{y_1}\mu_{w_2}}+\frac{\mathbb{E}[(\theta_{12}\bm{Y}_1-\bm{Z}_1)(\theta_{22}\bm{Y}_2-\bm{Z}_2)]}{\mu_{y_1}\mu_{y_2}})\notag\\
	&=\frac{1}{n}\mathbb{E}\left[(\frac{\theta_{11}\bm{W}_1-\bm{X}_1}{\mu_{w_1}}+\frac{\theta_{12}\bm{Y}_1-\bm{Z}_1}{\mu_{y_1}})(\frac{\theta_{21}\bm{W}_2-\bm{X}_2}{\mu_{w_2}}+\frac{\theta_{22}\bm{Y}_2-\bm{Z}_2}{\mu_{y_2}})\right].
\end{align}
If we add the subscript $i$, replace the subscript of $1$ by $t_1$ and $2$ by $t_2$, and let $\theta_{11}=\nu_{i,t_1}$, $\theta_{12}=\omega_{i,t_1}$, $\theta_{21}=\nu_{i,t_2}$, $\theta_{22}=\omega_{i,t_2}$, then
\begin{align}
	&Cov[\hat{V}_{i,t_1},\hat{V}_{i,t_2}]\notag\\
	&=\frac{1}{n_i}\mathbb{E}\left[(\frac{\nu_{i,t_1}\bm{W}_{i,t_1}-\bm{X}_{i,t_1}}{\mu_{w_{i,t_1}}}+\frac{\omega_{i,t_1}\bm{Y}_{i,t_1}-\bm{Z}_{i,t_1}}{\mu_{y_{i,t_1}}})(\frac{\nu_{i,t_2}\bm{W}_{i,t_2}-\bm{X}_{i,t_2}}{\mu_{w_{i,t_2}}}+\frac{\omega_{i,t_2}\bm{Y}_{i,t_2}-\bm{Z}_{i,t_2}}{\mu_{y_{i,t_2}}})\right].
\end{align}
If we use samples to estimate the covariance, the estimator will be
\begin{align}
	Cov_{i,t_1,t_2}=\sum_{j=1}^{n_i}\left(\frac{\hat{X}_{i,j,t_1}-\hat{\nu}_{i,t_1}\hat{W}_{i,j,t_1}}{\sum_{k=1}^{n_i}\hat{W}_{i,k,t_1}}+\frac{\hat{Z}_{i,j,t_1}-\hat{\omega}_{i,t_1}\hat{Y}_{i,j,t_1}}{\sum_{k=1}^{n_i}\hat{Y}_{i,k,t_1}}\right)\notag\\
	\quad \left(\frac{\hat{X}_{i,j,t_2}-\hat{\nu}_{i,t_2}\hat{W}_{i,j,t_2}}{\sum_{k=1}^{n_i}\hat{W}_{i,k,t_2}}+\frac{\hat{Z}_{i,j,t_2}-\hat{\omega}_{i,t_2}\hat{Y}_{i,j,t_2}}{\sum_{k=1}^{n_i}\hat{Y}_{i,k,t_2}}\right).
\end{align}
Using the same technique as \ref{ap:vsw}, we can prove that $n_i*Cov_{i,t_1,t_2}$ is strongly consistent for $E_{i,t_1,t_2}=\mathbb{E}\left[(\frac{\nu_{i,t_1}\bm{W}_{i,t_1}-\bm{X}_{i,t_1}}{\mu_{w_{i,t_1}}}+\frac{\omega_{i,t_1}\bm{Y}_{i,t_1}-\bm{Z}_{i,t_1}}{\mu_{y_{i,t_1}}})(\frac{\nu_{i,t_2}\bm{W}_{i,t_2}-\bm{X}_{i,t_2}}{\mu_{w_{i,t_2}}}+\frac{\omega_{i,t_2}\bm{Y}_{i,t_2}-\bm{Z}_{i,t_2}}{\mu_{y_{i,t_2}}})\right]$. 

\section{Formulations for the Estimators}
\subsection{General Formulations for the Estimators}
\label{ap:ce}

The naive mixture estimators for the four methods are
\begin{align}
	\hat{V}_{NMIS}&=\sum_{i=1}^M\alpha_i\hat{V}_{IS,i}\\
	\hat{V}_{NMDR}&=\sum_{i=1}^M\alpha_i\hat{V}_{DR,i}\\
	\hat{V}_{NMWIS}&=\sum_{i=1}^M\alpha_i\hat{V}_{SWIS,i}\\
	\hat{V}_{NMWDR}&=\sum_{i=1}^M\alpha_i\hat{V}_{SWDR,i}
\end{align}
After taking $t$ into account, the mixture estimators for the four methods are
\begin{align}
	\hat{V}_{MIS}&=\sum_{i=1}^M\sum_{t=0}^T\alpha_{i,t}\hat{V}_{IS,i,t}\\
	\hat{V}_{MDR}&=\sum_{i=1}^M\sum_{t=0}^T\alpha_{i,t}\hat{V}_{DR,i,t}\\
	\hat{V}_{MWIS}&=\sum_{i=1}^M\sum_{t=0}^T\alpha_{i,t}\hat{V}_{SWIS,i,t}\\
	\hat{V}_{MWDR}&=\sum_{i=1}^M\sum_{t=0}^T\alpha_{i,t}\hat{V}_{SWDR,i,t}
\end{align}
After splitting the control variates from DR and SWDR, the $\alpha\beta$ mixture estimators are 
\begin{align}
	\hat{V}_{MDR}&=\sum_{i=1}^M\sum_{t=0}^T(\alpha_{i,t}\hat{V}_{IS,i,t}+\beta_{i,t}\hat{W}_{DR,i,t})\\
	\hat{V}_{MWDR}&=\sum_{i=1}^M\sum_{t=0}^T(\alpha_{i,t}\hat{V}_{SWIS,i,t}+\beta_{i,t}\hat{W}_{SWDR,i,t})
\end{align}
\subsection{Components of the Estimators}
The sub-estimators in Appendix \ref{ap:ce} are listed below:
\begin{table}[H]
  \begin{center}
    \caption{Formulas for the components of the estimators.}
    \label{tab:tableee}
    \begin{tabular}{c|c|c} 
          Method&Target&Formulation\\
      \hline
      IS & $\hat{V}_{IS,i}$&$\frac{1}{n_i}\sum_{j=1}^{n_i}\sum_{t=0}^T\gamma^t\rho_{i,j,t}r_{i,j,t}$\\
      IS & $\hat{V}_{IS,i,t}$&$\frac{1}{n_i}\sum_{j=1}^{n_i}\gamma^t\rho_{i,j,t}r_{i,j,t}$\\
      DR & $\hat{V}_{DR,i}$&$\frac{1}{n_i}\sum_{j=1}^{n_i}\sum_{t=0}^T\gamma^t\left(\rho_{i,j,t-1}\hat{V}(s_{i,j,t})+\rho_{i,j,t}(r_{i,j,t}-\hat{Q}(s_{i,j,t},a_{i,j,t}))\right)$\\
      DR & $\hat{V}_{DR,i,t}$&$\frac{1}{n_i}\sum_{j=1}^{n_i}\gamma^t\left(\rho_{i,j,t-1}\hat{V}(s_{i,j,t})+\rho_{i,j,t}(r_{i,j,t}-\hat{Q}(s_{i,j,t},a_{i,j,t}))\right)$\\
      DR & $\hat{W}_{DR,i,t}$&$\frac{1}{n_i}\sum_{j=1}^{n_i}\gamma^t\left(\rho_{i,j,t-1}\hat{V}(s_{i,j,t})-\rho_{i,j,t}\hat{Q}(s_{i,j,t},a_{i,j,t})\right)$\\
      SWIS & $\hat{V}_{SWIS,i}$&$\sum_{j=1}^{n_i}\sum_{t=0}^T\gamma^tu_{i,j,t}r_{i,j,t}$\\
      SWIS & $\hat{V}_{SWIS,i,t}$&$\sum_{j=1}^{n_i}\gamma^tu_{i,j,t}r_{i,j,t}$\\
      SWDR & $\hat{V}_{SWDR,i}$&$\sum_{j=1}^{n_i}\sum_{t=0}^T\gamma^t\left(u_{i,j,t-1}\hat{V}(s_{i,j,t})+u_{i,j,t}(r_{i,j,t}-\hat{Q}(s_{i,j,t},a_{i,j,t}))\right)$\\
      SWDR & $\hat{V}_{SWDR,i,t}$&$\sum_{j=1}^{n_i}\gamma^t\left(u_{i,j,t-1}\hat{V}(s_{i,j,t})+u_{i,j,t}(r_{i,j,t}-\hat{Q}(s_{i,j,t},a_{i,j,t}))\right)$\\
      SWDR & $\hat{W}_{SWDR,i,t}$&$\sum_{j=1}^{n_i}\gamma^t\left(u_{i,j,t-1}\hat{V}(s_{i,j,t})-u_{i,j,t}\hat{Q}(s_{i,j,t},a_{i,j,t})\right)$
     \end{tabular}
  \end{center}
\end{table}

\subsection{Variance / Covariance Estimators for the Components}
\label{ap:ce3}
We can directly estimate the variances / covariances of components of IS and DR. In addition, with the formulas in Appendix \ref{ap:vsw} and \ref{ap:csw}, we can obtain the variance / covariance estimators for the components of SWIS and SWDR. 

For naive mixture estimators, we need to estimate $\mathbb{V}[\hat{V}_{IS,i}]$, $\mathbb{V}[\hat{V}_{DR,i}]$, $\mathbb{V}[\hat{V}_{SWIS,i}]$ and $\mathbb{V}[\hat{V}_{SWDR,i}]$. They are formulated as
\begin{align}
	n_i\mathbb{V}[\hat{V}_{IS,i}]&\approx\frac{1}{n_i}\sum_{j=1}^{n_i}\left(\sum_{t=0}^T\gamma^t\rho_{i,j,t}r_{i,j,t}\right)^2-\left(\frac{1}{n_i}\sum_{j=1}^{n_i}\sum_{t=0}^T\gamma^t\rho_{i,j,t}r_{i,j,t}\right)^2\\
	n_i\mathbb{V}[\hat{V}_{DR,i}]&\approx\frac{1}{n_i}\sum_{j=1}^{n_i}\left(\sum_{t=0}^T\gamma^t\left(\rho_{i,j,t-1}\hat{V}(s_{i,j,t})+\rho_{i,j,t}(r_{i,j,t}-\hat{Q}(s_{i,j,t},a_{i,j,t}))\right)\right)^2\notag\\
	&\quad -\left(\frac{1}{n_i}\sum_{j=1}^{n_i}\sum_{t=0}^T\gamma^t\left(\rho_{i,j,t-1}\hat{V}(s_{i,j,t})+\rho_{i,j,t}(r_{i,j,t}-\hat{Q}(s_{i,j,t},a_{i,j,t}))\right)\right)^2\\
	n_i\mathbb{V}[\hat{V}_{SWIS,i}]&\approx n_i\sum_{j=1}^{n_i}\left(\sum_{t=0}^Tu_{i,j,t}\left(\gamma^tr_{i,j,t}-\hat{\theta}_{i,t}\right)\right)^2\\
	n_i\mathbb{V}[\hat{V}_{SWDR,i}]&\approx n_i\sum_{j=1}^{n_i}\Bigg(\sum_{t=0}^Tu_{i,j,t}\left(\gamma^t\left(r_{i,j,t}-\hat{Q}(s_{i,j,t},a_{i,j,t})\right)-\hat{\nu}_{i,t}\right)\notag\\
	&\quad + \sum_{t=0}^Tu_{i,j,t-1}\left(\gamma^t\hat{V}(s_{i,j,t})-\hat{\omega}_{i,t}\right)\Bigg)^2
\end{align}
where $\hat{\theta}_{i,t}=\sum_{j=1}^{n_i}\gamma^tu_{i,j,t}r_{i,j,t}$, $\hat{\nu}_{i,t}=\sum_{j=1}^{n_i}\gamma^tu_{i,j,t}\left(r_{i,j,t}-\hat{Q}(s_{i,j,t},a_{i,j,t})\right)$ and $\hat{\omega}_{i,t}=\sum_{j=1}^{n_i}\gamma^tu_{i,j,t-1}\hat{V}(s_{i,j,t})$.

For mixture estimators, we will need to estimate $Cov[\hat{V}_{IS,i,t_1},\hat{V}_{IS,i,t_2}]$, $Cov[\hat{V}_{DR,i,t_1},\hat{V}_{DR,i,t_2}]$, $Cov[\hat{V}_{SWIS,i,t_1},\hat{V}_{SWIS,i,t_2}]$ and $Cov[\hat{V}_{SWDR,i,t_1},\hat{V}_{SWDR,i,t_2}]$. Their formulations are
\begin{align}
	n_iCov[\hat{V}_{IS,i,t_1},\hat{V}_{IS,i,t_2}]&\approx\frac{1}{n_i}\sum_{j=1}^{n_i}\gamma^{t_1+t_2}\rho_{i,j,t_1}\rho_{i,j,t_2}r_{i,j,t_1}r_{i,j,t_2}\notag\\
	&\quad -\left(\frac{1}{n_i}\sum_{j=1}^{n_i}\gamma^{t_1}\rho_{i,j,t_1}r_{i,j,t_1}\right)\left(\frac{1}{n_i}\sum_{j=1}^{n_i}\gamma^{t_2}\rho_{i,j,t_2}r_{i,j,t_2}\right)\\
	n_iCov[\hat{V}_{DR,i,t_1},\hat{V}_{DR,i,t_2}]&\approx\frac{1}{n_i}\sum_{j=1}^{n_i}\left(\gamma^{t_1}\left(\rho_{i,j,t_1-1}\hat{V}(s_{i,j,t_1})+\rho_{i,j,t_1}(r_{i,j,t_1}-\hat{Q}(s_{i,j,t_1},a_{i,j,t_1}))\right)\right)\notag\\
	&\quad *\left(\gamma^{t_2}\left(\rho_{i,j,t_2-1}\hat{V}(s_{i,j,t_2})+\rho_{i,j,t_2}(r_{i,j,t_2}-\hat{Q}(s_{i,j,t_2},a_{i,j,t_2}))\right)\right)\notag\\
	&\quad -\left(\frac{1}{n_i}\sum_{j=1}^{n_i}\gamma^{t_1}\left(\rho_{i,j,t_1-1}\hat{V}(s_{i,j,t_1})+\rho_{i,j,t_1}(r_{i,j,t_1}-\hat{Q}(s_{i,j,t_1},a_{i,j,t_1}))\right)\right)\notag\\
	&\quad *\left(\frac{1}{n_i}\sum_{j=1}^{n_i}\gamma^{t_2}\left(\rho_{i,j,t_2-1}\hat{V}(s_{i,j,t_2})+\rho_{i,j,t_2}(r_{i,j,t_2}-\hat{Q}(s_{i,j,t_2},a_{i,j,t_2}))\right)\right)\\
		n_iCov[\hat{V}_{SWIS,i,t_1},\hat{V}_{SWIS,i,t_2}]&\approx n_i\sum_{j=1}^{n_i}u_{i,j,t_1}u_{i,j,t_2}\left(\gamma^{t_1}r_{i,j,t_1}-\hat{\theta}_{i,t_1}\right)\left(\gamma^{t_2}r_{i,j,t_2}-\hat{\theta}_{i,t_2}\right)
	\end{align}
	\begin{align}
	n_iCov[\hat{V}_{SWDR,i,t_1},\hat{V}_{SWDR,i,t_2}]&\approx n_i\sum_{j=1}^{n_i}\Bigg(u_{i,j,t_1}\left(\gamma^{t_1}\left(r_{i,j,t_1}-\hat{Q}(s_{i,j,t_1},a_{i,j,t_1})\right)-\hat{\nu}_{i,t_1}\right)\notag\\
	&\quad + u_{i,j,t_1-1}\left(\gamma^{t_1}\hat{V}(s_{i,j,t_1})-\hat{\omega}_{i,t_1}\right)\Bigg)\notag\\
	&\quad *\Bigg(u_{i,j,t_2}\left(\gamma^{t_2}\left(r_{i,j,t_2}-\hat{Q}(s_{i,j,t_2},a_{i,j,t_2})\right)-\hat{\nu}_{i,t_2}\right)\notag\\
	&\quad + u_{i,j,t_2-1}\left(\gamma^{t_2}\hat{V}(s_{i,j,t_2})-\hat{\omega}_{i,t_2}\right)\Bigg)
\end{align}
For $\alpha\beta$ mixture estimators, we need to estimate $Cov[\hat{V}_{IS,i,t_1},\hat{W}_{DR,i,t_2}]$, $Cov[\hat{W}_{DR,i,t_1},\hat{W}_{DR,i,t_2}]$, $Cov[\hat{V}_{SWIS,i,t_1},\hat{W}_{SWDR,i,t_2}]$ and $Cov[\hat{W}_{SWDR,i,t_1},\hat{W}_{SWDR,i,t_2}]$. They can be approximated by
\begin{align}
	n_iCov[\hat{V}_{IS,i,t_1},\hat{W}_{DR,i,t_2}]&\approx \frac{1}{n_i}\sum_{j=1}^{n_i}\gamma^{t_1+t_2}\rho_{i,j,t_1}r_{i,j,t_1}\left(\rho_{i,j,t_2-1}\hat{V}(s_{i,j,t_2})-\rho_{i,j,t_2}\hat{Q}(s_{i,j,t_2},a_{i,j,t_2})\right)\notag\\
	&\quad -\left(\frac{1}{n_i}\sum_{j=1}^{n_i}\gamma^{t_1}\rho_{i,j,t_1}r_{i,j,t_1}\right)\notag\\
	&\quad *\left(\frac{1}{n_i}\sum_{j=1}^{n_i}\gamma^{t_2}\left(\rho_{i,j,t_2-1}\hat{V}(s_{i,j,t_2})-\rho_{i,j,t_2}\hat{Q}(s_{i,j,t_2},a_{i,j,t_2})\right)\right)\\
	n_iCov[\hat{W}_{DR,i,t_1},\hat{W}_{DR,i,t_2}]&\approx \frac{1}{n_i}\sum_{j=1}^{n_i}\gamma^{t_1+t_2}\left(\rho_{i,j,t_1-1}\hat{V}(s_{i,j,t_1})-\rho_{i,j,t_1}\hat{Q}(s_{i,j,t_1},a_{i,j,t_1})\right)\notag\\
	&\quad * \left(\rho_{i,j,t_2-1}\hat{V}(s_{i,j,t_2})-\rho_{i,j,t_2}\hat{Q}(s_{i,j,t_2},a_{i,j,t_2})\right)\notag\\
	&\quad -\left(\frac{1}{n_i}\sum_{j=1}^{n_i}\gamma^{t_1}\left(\rho_{i,j,t_1-1}\hat{V}(s_{i,j,t_1})-\rho_{i,j,t_1}\hat{Q}(s_{i,j,t_1},a_{i,j,t_1})\right)\right)\notag\\
	&\quad *\left(\frac{1}{n_i}\sum_{j=1}^{n_i}\gamma^{t_2}\left(\rho_{i,j,t_2-1}\hat{V}(s_{i,j,t_2})-\rho_{i,j,t_2}\hat{Q}(s_{i,j,t_2},a_{i,j,t_2})\right)\right)\\
	n_iCov[\hat{V}_{SWIS,i,t_1},\hat{W}_{SWDR,i,t_2}]&\approx n_i\sum_{j=1}^{n_i}u_{i,j,t_1}\left(\gamma^{t_1}r_{i,j,t_1}-\hat{\theta}_{i,t_1}\right)\notag\\
	&\quad *\left(u_{i,j,t_2-1}\left(\gamma^{t_2}\hat{V}(s_{i,j,t_2})-\hat{\phi}_{i,t_2}\right)-u_{i,j,t_2}\left(\gamma^{t_2}\hat{Q}(s_{i,j,t_2},a_{i,j,t_2})-\hat{\psi}_{i,t_2}\right)\right)\\
	n_iCov[\hat{W}_{SWDR,i,t_1},\hat{W}_{SWDR,i,t_2}]&\approx n_i\sum_{j=1}^{n_i}\left(u_{i,j,t_1-1}\left(\gamma^{t_1}\hat{V}(s_{i,j,t_1})-\hat{\phi}_{i,t_1}\right)-u_{i,j,t_1}\left(\gamma^{t_1}\hat{Q}(s_{i,j,t_1},a_{i,j,t_1})-\hat{\psi}_{i,t_1}\right)\right)\notag\\
	&\quad *\left(u_{i,j,t_2-1}\left(\gamma^{t_2}\hat{V}(s_{i,j,t_2})-\hat{\phi}_{i,t_2}\right)-u_{i,j,t_2}\left(\gamma^{t_2}\hat{Q}(s_{i,j,t_2},a_{i,j,t_2})-\hat{\psi}_{i,t_2}\right)\right)
\end{align}
where $\phi_{i,t}=\sum_{j=1}^{n_i}\gamma^{t}u_{i,j,t}\hat{V}(s_{i,j,t})$ and $\psi_{i,t}=\sum_{j=1}^{n_i}\gamma^{t}u_{i,j,t}\hat{Q}(s_{i,j,t},a_{i,j,t})$.

Note that all of the above estimators multiplied by $n_i$ are strongly consistent for some value. Therefore, we can easily show that the three mixture estimators are all strongly consistent for $V$.
\section{Experiment Details}
\subsection{Environmental Settings}
\label{ap:es}
In this section, there are no random variables so we use bold letters to denote vectors.

In the recommendation environment, we first define the number of topics $T=100$. $1000$ documents are generated and each is observed by a $T$ dimension vector. Each document should reflect whether it is related to each topic and each user should indicate the preference of each topic. For the topics, we generate an abundance vector $\bm{A}=[A_1,A_2,...,A_T]^T$ and a quality vector $\bm{Q}=[Q_1,Q_2,...,Q_T]^T$. The abundance vector satisfies $\sum_{i=1}^TA_i=1$ and $\forall i$, $A_i\ge 0$. When we generate a relevant topic for a document, the abundance vector specifies the probability of generating each topic. The quality vector satisfies $\forall i$, $0\le Q_i\le 1$, indicating the original quality of each topic. 

For each document, we define relevance vector $\bm{r}_i\in\{0,1\}^T$ and quality $q_i$. When the document sampler is generating the $i$-th document, it will first generate three topics $t_1$, $t_2$ and $t_3$ (repeatable) by $\bm{A}$, and set the $t_1$-th, $t_2$-th and $t_3$-th dimensions of $\bm{r}_i$ to 1 while setting the other dimensions to 0. The quality is calculated by $q_i=(\bm{Q}^T\bm{r}_i+U)/2$, where $U$ is generated uniformly from $[0,1]$. To simulate the real environment, only $\bm{r}_i$ of each document can be observed. 

At each time, the hidden state consists of
\begin{itemize}
	\item user id $i$, indicating the current user;
	\item interest $I$, determining the satisfaction of the user;
	\item satisfaction $s$, influencing the reward of policy;
	\item document vector $\bm{d}$, the relevance vector of the current document.
\end{itemize}
We generated 5 preference vectors $\bm{p}_i\in[-1,1]^T$. Each of them represents one user. When the user sampler generates the initial state, it randomly pick one from the 5 users, generate an initial interest $I$, and set the satisfaction by $s=\frac{1}{1+e^{-0.5I}}$. The initial document vector is set as $[1,1,...,1]^T$. To simulate the real environment, only $i$ and $\bm{d}$ can be observed by the agent.

Now we define the reward function and the state transition function. Suppose we recommend the $j$-th document to the $i$-th user. We define the liking of the user to the document by $l_{i,j}=\overrightarrow{e}^T(\bm{r}_j\circ\bm{p}_i\circ(\bm{d}+0.5)/2)$, where $\circ$ is element-wise multiplication. This formula indicates the expectation of the user about the next document. It should be close to his or her preference as well as the current document. The probability of taking the document is $\frac{l_{i,j}}{1+l_{i,j}}$ while the probability of leaving is $\frac{1}{1+l_{i,j}}$. If the user takes the document, we will calculate its engagement time by $t_{i,j}\sim\mathcal{N}(q_j,0.1^2)$ and produce the reward of $s*e^{t_{i,j}}$. The reward depends on the quality of the document as well as the user satisfaction. 

After the choice of user, the state vector will change. The interest will be updated by $I'=0.9I+\bm{p}_i^T\bm{r}_j+V$, where $V\sim\mathcal{N}(0,0.1^2)$. Meanwhile, the document vector is updated by $\bm{d}_i'=\bm{r}_j$. Note we also need to update the interest by $s'=\frac{1}{1+e^{-0.5I'}}$.
\subsection{Implementation of REINFORCE}
\label{ap:ir}
In the above environment, at each step, the agent needs to propose a document only with $i$, $\bm{d}$ and document list $[\bm{r}_1,\bm{r}_2,...,\bm{r}_D]$. We can train a policy network to represent the policy $\pi(a|s)$. However, when the list is long(action space is large), it is hard to directly output the policy $\pi(a|s)$ from neural networks. To exploit the structure of documents, we reduce the output of the policy network to a T dimension vector $\bm{y}$ and compute the policy by $\pi(a_i|s)\propto \bm{y}^T\bm{r}_i$. This not only solves the above problem but also reduces the time complexity of sampling. We initialize $\bm{b}_i=\sum_{j=1}^i\bm{r}_j$. When sampling, we can generate a number $c$ from $[0,1]$ and use binary search to find the smallest k satisfying $\bm{y}^T\bm{b}_k\ge c$. The $k$-th document would then be the sampled document. The time complexity is $O(TlogD)$, which is useful when the document list is long.

\subsection{Implementation of OPE Algorithms}
\label{ap:id}

In the same environment, we train 100 different REINFORCE policies $p_1,p_2,...,p_{100}$. Each policy collects 1000000 offline data, with the $i$-th defined by $DATA_i$. In our experiments, the policies are set up as target policy and behavior policies in turn. Specifically, for the $i$-th experiment, we set the target policy as $p_{i}$ and use $DATA_i$ to estimate the true mean reward $\overline{\theta}_i$. The behavior policies are set as $\forall j\in\{1,2,...,M\}\ \pi_j=p_{(i+j-1)\%100+1}$ and the offline data are $(DATA_{i+1},DATA_{i+2},...,DATA_{(i+j-1)\%100+1})$. With the settings, the $i$-th evaluation result is $\hat{\theta}_i$ and the squared error is computed by $(\hat{\theta}_i-\overline{\theta}_i)^2$. We use the first 50 experiments as validation set to tune parameters and the last 50 experiments as test set to report the results.

We trained DM with 10000 data. The reward function $\mathbb{E}[R(s,a)]$ is approximated by Bayesian Ridge Regressor \cite{scikit-learn}. The transition function $P(\cdot|s,a)$ is approximated by neural network. The network consists of 2 fully connected layers with 512 units and an output layer with 2 units. The hidden layers are activated by relu and the last layer is activated by softmax. We minimize the cross entropy loss function with Adam optimizer \cite{Kingma2015AdamAM}. The batch size is 32 and we train for 600 epochs. For the first 300 epochs the learning rate is set as 0.0001. For the last 300 epochs the learning rate is 0.00001. After approximating the environment functions, we iterate by formula (\ref{iteration}) for 20 times and get the estimated value functions $\hat{Q}(s,a)$ and $\hat{V}(s)$. 

To reduce the variance of the methods we clip the importance ratios. For IS based methods we replace $\rho_{i,j,t}$ by $\overline{\rho_{i,j,t}}=\min(\rho_{i,j,t},2000)$. For DR based methods we make $\hat{V}_{i,j,t}=\gamma^t(\overline{\rho_{i,j,t-1}}\frac{\pi(a_{i,j,t}|s_{i,j,t})}{\pi_i(a_{i,j,t}|s_{i,j,t})}(r_{i,j,t}-\hat{Q}(s_{i,j,t},a_{i,j,t}))+\overline{\rho_{i,j,t-1}}\hat{V}(s_{i,j,t}))$. Note that clipping introduces additional bias. So our methods can be further improved by considering the bias.

\subsection{Tuning of Hyperparameter T}
\label{ap:ih}
\begin{figure}
\begin{minipage}{0.5\textwidth}
  \centering
  \includegraphics[width=.95\linewidth]{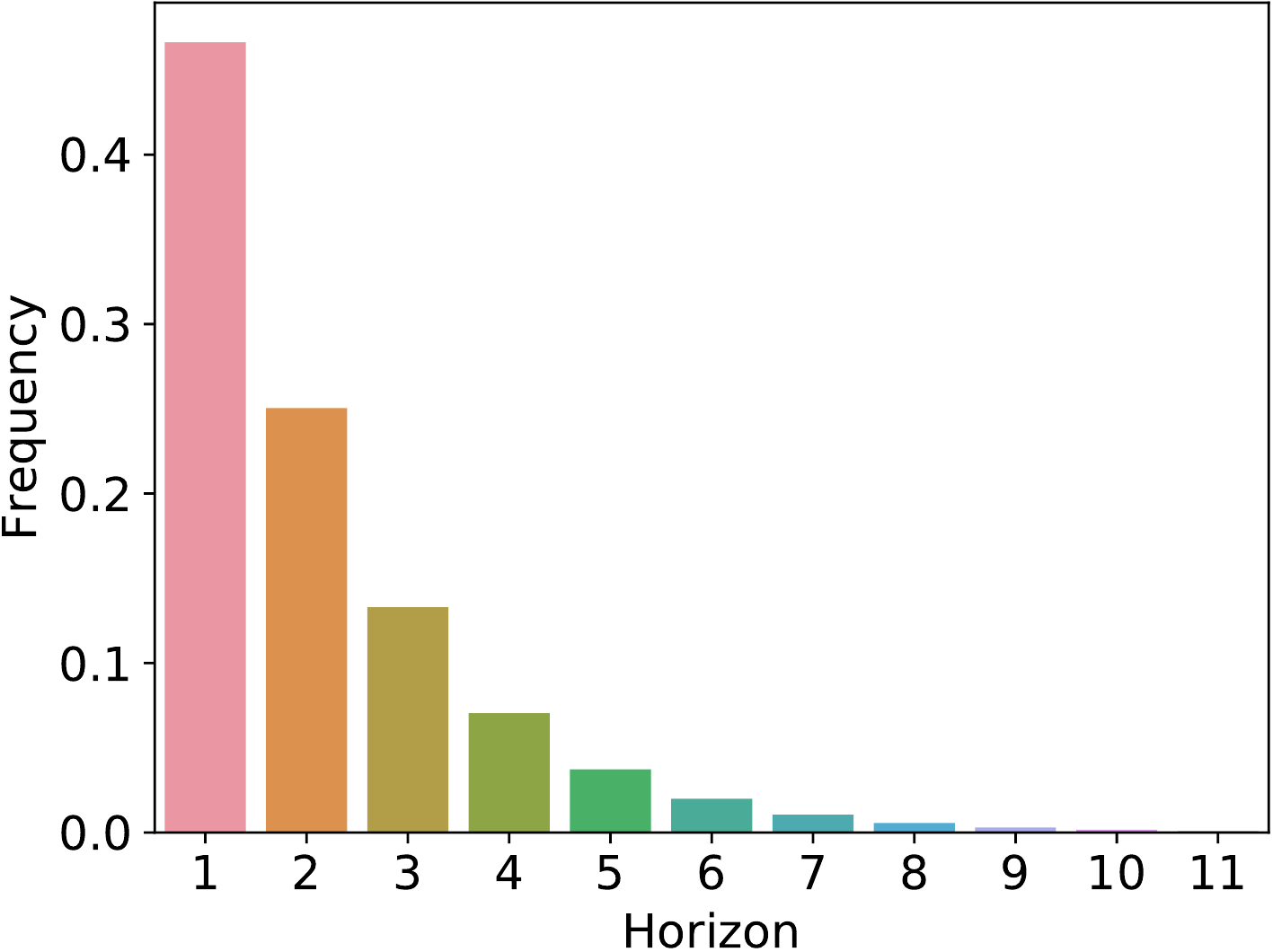}
  \caption{Distribution of length of data.}
  \label{fig:datalength}
\end{minipage}\hfill
\begin{minipage}{0.5\textwidth}
  \centering
  \includegraphics[width=.95\linewidth]{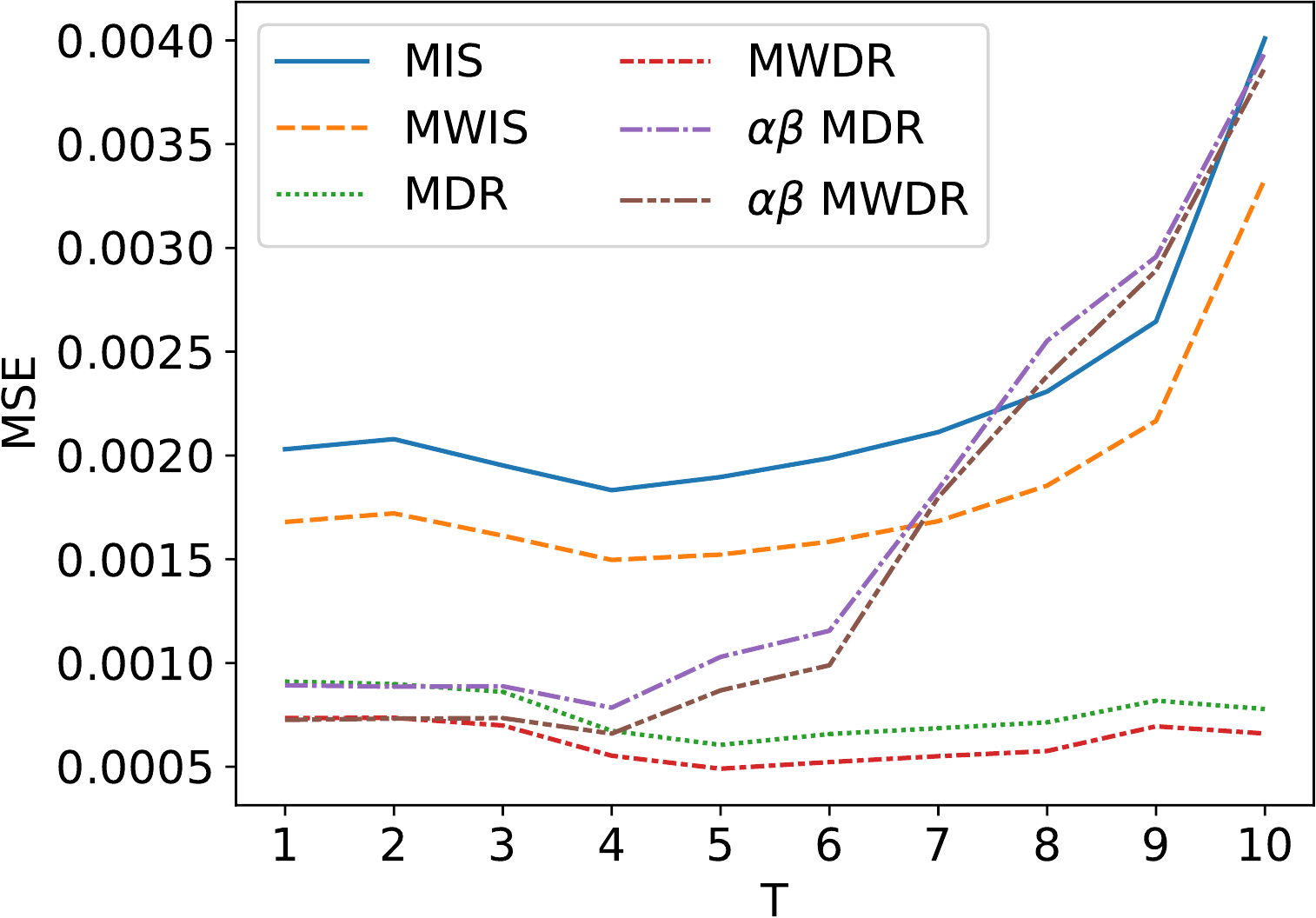}
  \caption{MSE of the methods with different T.}
  \label{fig:mseatt}
\end{minipage}
\end{figure}
In mixture estimators and $\alpha\beta$ mixture estimators, we choose a hyper-parameter $T$, mix the values from 0 to T and simply add up the remains. This is because the length of data from each behavior policy is random, as Figure \ref{fig:datalength}. When $t$ is large, the reward decreases exponentially and the variances and covariances about $\hat{V}_{i,t}$ also decreases, making the matrixes nearly singular. Such problem leads to the phenomenon in Figure \ref{fig:mseatt}. Numerical results can be found in Appendix \ref{ap:at}. When T is small, the MSE decreases as T increases because more values are mixed. When T is large, the MSE increases as T increases because of the amplification of error from ill covariance matrixes. By Figure \ref{fig:mseatt}, we set T=4 for MIS, MWIS, $\alpha\beta$ MDR and $\alpha\beta$ MWDR and set T=5 for MDR and MWDR. 

\section{Numerical Results}
\subsection{MSE of different methods with different T}
\label{ap:at}
\centering
    \begin{tabularx}{0.83\linewidth}{ccccccc}
      \toprule[1.5pt]
       T &  MIS &  MWIS &  MDR&  MWDR&  $\alpha\beta$ MDR&  $\alpha\beta$ MWDR \\\midrule[1pt]
1 & 0.00203 & 0.001679 & 0.00091 & 0.000734 & 0.000893 & 0.000726 \\
2 & 0.002079 & 0.001721 & 0.000898 & 0.000736 & 0.000886 & 0.000732 \\
3 & 0.001952 & 0.001613 & 0.000861 & 0.000699 & 0.000888 & 0.000735 \\
4 & \textbf{0.001833} & \textbf{0.001497} & 0.000673 & 0.000553 & \textbf{0.000785} & \textbf{0.00066} \\
5 & 0.001896 & 0.001522 & \textbf{0.000605} & \textbf{0.000491} & 0.001029 & 0.000868 \\
6 & 0.001987 & 0.001584 & 0.000658 & 0.000522 & 0.001155 & 0.000989 \\
7 & 0.002113 & 0.001684 & 0.000686 & 0.000551 & 0.001838 & 0.001798 \\
8 & 0.002308 & 0.001855 & 0.000714 & 0.000576 & 0.002552 & 0.002383 \\
9 & 0.002646 & 0.002166 & 0.000818 & 0.000695 & 0.002957 & 0.002891 \\
10 & 0.004009 & 0.003332 & 0.000778 & 0.00066 & 0.003941 & 0.003869 \\
      \bottomrule[1.5pt]
    \end{tabularx}
\subsection{MSE of different methods with different M}
\label{ap:am}
\centering

    \begin{tabularx}{0.77\linewidth}{cccccc}
      \toprule[1.5pt]
       M & 1 & 2 & 3&  4&  5 \\\midrule[1pt]
IS& 0.001877& 0.001703& 0.001544& 0.001515& 0.001344\\
WIS& 0.001525& 0.001351& 0.00123& 0.001217& 0.001075\\
SWIS& 0.001525& 0.001367& 0.001252& 0.001241& 0.001099\\
NMIS& 0.001942& 0.001435& 0.001289& 0.001158& 0.001017\\
NMWIS& 0.001576& 0.001139& 0.001041& 0.000953& 0.000836\\
MIS& 0.001907& 0.001452& 0.001394& 0.0013& 0.001126\\
MWIS& 0.001533& 0.001152& 0.001126& 0.001075& 0.000928\\
DR& 0.00082& 0.000455& 0.000309& 0.000381& 0.000377\\
WDR& \textbf{0.000675}& 0.000357& \textbf{0.000235}& 0.0003& 0.000301\\
SWDR& \textbf{0.000675}& 0.000352& \textbf{0.000235}& 0.0003& 0.000299\\
NMDR& 0.000968& 0.000422& 0.000284& 0.0004& 0.000395\\
NMWDR& 0.000883& 0.000375& 0.000253& 0.000337& 0.000333\\
MDR& 0.000861& 0.000398& 0.000294& 0.000264& 0.000311\\
MWDR& 0.000776& \textbf{0.000334}& 0.000247& \textbf{0.00021}& \textbf{0.000245}\\
$\alpha\beta$MDR& 0.001017& 0.000469& 0.00041& 0.000408& 0.000371\\
$\alpha\beta$MWDR& 0.000911& 0.000394& 0.000349& 0.000344& 0.000317\\
      \bottomrule[1.5pt]
    \end{tabularx}

\subsection{MSE and condition number of different methods}
\label{ap:mcd}
\centering

    \begin{tabularx}{0.7\linewidth}{ccc}
      \toprule[1.5pt]
      Method & MSE &  Cond Number \\\midrule[1pt]
IS& 0.0013444257969445908 &/\\
WIS& 0.0010745378682455794 &/\\
SWIS& 0.001099086646547933 &/\\
NMIS& 0.0010165986763763788 &/\\
NMWIS& 0.0008363384188827019 &/\\
MIS& 0.0011259428162672161 & 15.233859883668414 \\
MWIS& 0.0009282956141402447 & 17.60016020307187 \\
DR& 0.00037712737924285305 &/\\
WDR& 0.0003011794977270172 &/\\
SWDR& 0.00029893021636011546 &/\\
NMDR& 0.0003947548680215243 &/\\
NMWDR& 0.00033273193073131153 &/\\
MDR& 0.0003109721026979073 & 51.790423695019456 \\
MWDR& \textbf{0.0002449612080194226} & 74.18300868487523 \\
$\alpha\beta$MDR& 0.00037141793259772417 & 320.712344891526 \\
$\alpha\beta$MWDR& 0.0003171837312053011 & 337.9175903029748 \\
      \bottomrule[1.5pt]
    \end{tabularx}

\end{document}